\documentclass[10pt,twocolumn,letterpaper]{article}
\usepackage[pagenumbers]{cvpr}
\ifdefined\XeTeXrevision
  \usepackage{fontspec}
\fi
\usepackage{graphicx}
\usepackage[font=small]{caption}
\usepackage{amsmath,amssymb}
\usepackage{booktabs}
\usepackage{multirow}
\usepackage{xcolor}
\usepackage{xspace}
\usepackage{tikz}
\usetikzlibrary{positioning,arrows.meta,fit,calc,shapes.misc,decorations.pathreplacing}
\usepackage{algorithm}
\usepackage[noend]{algpseudocode}
\usepackage{microtype}
\usepackage{siunitx}
\newcommand{\vggto}{VGGT-$\Omega$\xspace}

\newcommand{\x}{$\times$\xspace}

\newcommand{\ra}{RAWATTA\xspace}
\newcommand{\raf}{RAWATTA-Fast\xspace}

\definecolor{cvprblue}{rgb}{0.21,0.49,0.74}
\usepackage[pagebackref,breaklinks,colorlinks,allcolors=cvprblue]{hyperref}

\def\confName{CVPR}
\def\confYear{2027}

\title{Training-Free Acceleration of Multi-View 3D Transformers\\via Regime-Aware Attention Approximation}
\author{M. Izzet Sezer\\
Independent Researcher\\
{\tt\small muhammedsezer12@gmail.com}
\and
Sevval Belkis Dikkaya\\
Carnegie Mellon University\\
{\tt\small sdikkaya@andrew.cmu.edu}
}

\begin{document}
\twocolumn[{\renewcommand\twocolumn[1][]{#1}\maketitle\vspace{-2.2em}
\begin{center}
\includegraphics[width=\linewidth]{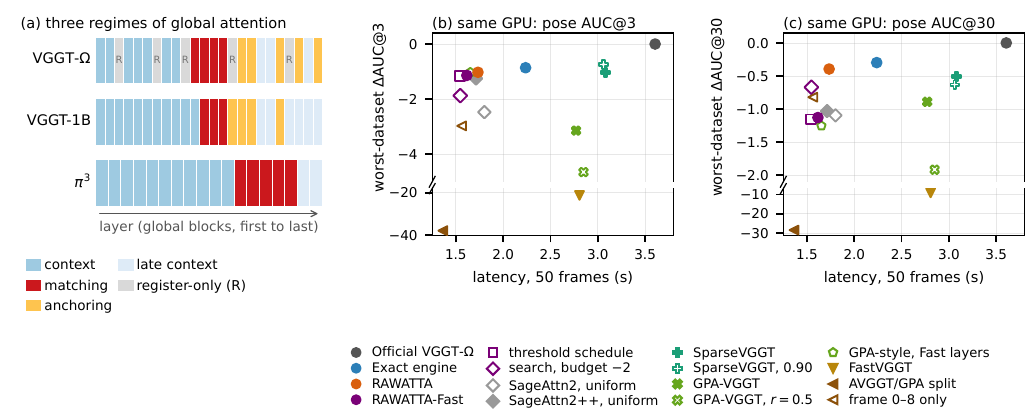}
\vspace{-1.9em}
\captionof{figure}{\textbf{(a)} Global attention in multi-view transformers follows three regimes (10 frames, i.e.\ images, 7 datasets; \cref{sec:analysis}); \vggto attends strongly to the first (reference) frame, VGGT-1B weakly, and $\pi^3$, which has no reference frame, not at all. \textbf{(b, c)} On one RTX 4070 SUPER: latency at 50 frames against the change in pose AUC@3 (b) and AUC@30 (c), measured on 10-frame sequences, on the worst of six datasets, relative to Official \vggto. Filled: \ra settings; hollow purple: other schedules; grey: uniform low-precision kernels; green/gold: published accelerators; brown: the AVGGT/GPA-VGGT layer split in our engine. Accelerators that merge or drop keys in matching layers lose by far the most accuracy.}
\label{fig:teaser}
\end{center}\vspace{0.6em}}]
\begin{abstract}
Multi-view 3D transformers such as \vggto reconstruct a scene from a set of unposed images in a single forward pass, predicting camera poses, depth and 3D points, but the cost of their global attention grows quadratically with the number of images. Accelerators borrowed from large language models assume that each token attends to only a few others; on these models they lose accuracy or gain little speed. We study the global attention of three multi-view transformers on seven datasets and find that every layer plays one of three roles: it averages over the whole scene, matches points across many images, or anchors on the first image, whose coordinates the outputs use. Building on this analysis, we propose Regime-AWare ATTention Approximation (\ra), a training-free method that runs attention in 8-bit precision and keeps 16-bit precision only in the layers where a worst-case calibration check requires it. Its fast setting, \raf, also approximates attention in a few averaging and anchoring layers, while matching layers and the first image keep all their keys. On a 12\,GB GPU at 50 images, \ra runs \vggto 2.1\x faster than the released code, mostly through our CUDA engine, and loses at most about one point of pose accuracy (AUC@3) on any dataset; \raf is slightly faster and less accurate. Both fit 200 images, where the released code runs out of memory at 100.
\end{abstract}

\section{Introduction}
\label{sec:intro}
Feed-forward multi-view transformers~\cite{vggt,pi3,vggtomega} reconstruct a scene from a set of photos in a single forward pass. Given unposed images, they predict where each camera was and where it pointed, a depth map for each image, and a 3D point for each pixel, which classic pipelines such as COLMAP~\cite{colmap} obtain by matching features and iterative optimisation. Their largest cost is global attention, in which every image patch attends to every patch of every other image, so the cost grows quadratically with the number of images. On a 12\,GB GPU, the released code~\cite{vggtomega} (Official \vggto) runs out of memory at 100 images.

Existing accelerators for these models reuse techniques developed for large language models: computing attention only for selected blocks of tokens (block-sparse attention~\cite{sparsevggt,saf3r}), merging similar tokens~\cite{fastvggt,gpavggt,httm}, letting some layers attend only within their own image~\cite{avggt}, and computing attention in low precision~\cite{sageattn2,quantvggt}. The first three assume that each token needs only a few other tokens, so the rest can be dropped or merged. On multi-view 3D models these methods either lose much accuracy or give little speed-up (\cref{fig:teaser}b).

We study where the global attention of three such models goes and find that it is not sparse but structured. Building on the layer roles reported by AVGGT~\cite{avggt} and RegimeVGGT~\cite{regimevggt}, we assign each global layer one of three \emph{regimes}. In \emph{diffuse context} layers, mostly early, each token spreads its attention almost evenly over all images, so the layer computes a scene-wide average. In \emph{selective matching} layers, in the middle, each token concentrates its attention, but on evidence spread over many images. In \emph{reference anchoring} layers, late in models that express their outputs in the first image's coordinates, every token sends about half of its attention to that first image. We call them context, matching and anchoring layers. The regimes are stable across seven datasets and 10 to 200 images, and appear in different arrangements in \vggto, VGGT-1B~\cite{vggt} and $\pi^3$~\cite{pi3} (\cref{fig:teaser}a). They explain why existing accelerators fail: there is little to drop where attention is spread evenly, and dropping or merging tokens destroys the correspondences that matching layers compute.

The regimes also tell which approximation each layer tolerates. A context layer computes an average over the scene, and an average changes little when each image is summarised by fewer keys: we merge every $4{\times}4$ group of patch keys into their mean or, in the earliest layers, let each image attend only to itself. An anchoring layer tolerates the same merging if the first image keeps all of its tokens, because that is where its attention concentrates. A matching layer needs every token and stays dense. We use this analysis to propose Regime-AWare ATTention Approximation (\ra), a training-free method for \vggto that runs attention in 8-bit precision and, in its fast setting (\raf), gives some context and anchoring layers one of these cheaper operators. Because errors of single layers do not simply add up, \ra accepts a complete configuration only if the worst of five calibration datasets stays within a fixed accuracy budget.

On an RTX 4070 SUPER at 50 images, \ra is 2.08\x faster than Official \vggto, 1.61\x of it from our CUDA engine, and at 10 and 50 images its pose accuracy drops by at most about one point of AUC@3 on every dataset (the area under the pose-accuracy curve up to a 3$^\circ$ error; AUC@30: up to 30$^\circ$; \cref{sec:exp}). \raf is 2.23\x faster but loses up to 2.2 points on one dataset at 50 images. At 200 images the method alone gives 1.7--2.0\x, and both settings fit on the card.

\noindent Our contributions are:
\begin{itemize}\setlength\itemsep{1pt}
\item \textbf{An analysis of global attention} in three multi-view transformers on seven datasets and 10--200 images. Extending the layer roles of AVGGT and RegimeVGGT, we assign a regime to every layer and identify \emph{reference anchoring}: whole layers that attend to the first image because of its position, and that are absent in $\pi^3$. The analysis explains why sparse attention, routing and token merging fail on these models (\cref{sec:analysis}).
\item \textbf{\ra}, a training-free method that sets a numeric format for each layer and an attention operator for each regime, and accepts a configuration only if the worst calibration dataset stays within an accuracy budget (\cref{sec:method}).
\item \textbf{A same-GPU comparison} with published accelerators that separates the gain of our engine from that of the method, with controls for which layers are approximated (\cref{sec:exp}).
\end{itemize}

\section{Related work}
\label{sec:related}
\paragraph{Feed-forward multi-view geometry.}
DUSt3R~\cite{dust3r} takes two images and predicts a 3D point for every pixel of both, in the first image's coordinates; a scene from many images then needs all pairs aligned by optimisation. MASt3R~\cite{mast3r} adds features for dense matching. VGGT~\cite{vggt} removes the pairwise step: it processes all images at once and predicts cameras, depth and point maps in the first image's coordinates, alternating attention within each image (frame attention) and across all images (global attention) over DINOv2~\cite{dinov2} patch tokens. $\pi^3$~\cite{pi3} predicts without a reference image, so its output does not depend on the image order. \vggto~\cite{vggtomega} scales the design and, as a \emph{training-time} measure, restricts five global layers to camera and register~\cite{registers} tokens. Long-sequence variants chunk the input~\cite{vggtlong,swiftvggt}, split it into subscenes~\cite{svggt}, or stream with a KV cache~\cite{streamvggt} bounded training-free~\cite{infinitevggt,evict3r,ovggt,regvggt,incvggt}. These methods change the training or split the sequence; we keep the released weights and the full sequence, and make each forward pass cheaper.

\paragraph{Accelerating VGGT-family models.}
\emph{Merging and pruning:} FastVGGT~\cite{fastvggt} merges tokens in global attention, Co-Me~\cite{comerge} merges low-confidence tokens with a distilled predictor, HTTM~\cite{httm} merges head-wise and PrePARE~\cite{prepare} prunes tokens. \emph{Sparse and structured attention:} SparseVGGT~\cite{sparsevggt} uses block-sparse global attention, HeSS~\cite{hess} adds head sensitivity, SAF3R~\cite{saf3r} uses head-wise dynamic sparsity, and FlashVGGT~\cite{flashvggt} trains descriptor-token cross-attention. AVGGT~\cite{avggt} first observed that early global layers of VGGT do not perform correspondence and converts blocks 0--8 to frame attention; GPA-VGGT~\cite{gpavggt} adds shared cross-frame keys there and merges keys in later blocks with a $\log s$ correction, and reports a \vggto port. Both fix contiguous layer ranges by inspection. TurboVGGT~\cite{turbovggt} trains adaptive sparse global attention. BlazeVGGT~\cite{blazevggt} (camera-token frame selection), VGGT-Prime~\cite{vggtprime} and Spark3R~\cite{spark3r} are concurrent and report their largest gains at 1,000--2,000 frames. These methods choose where to approximate by contiguous layer ranges or by per-token scores. We choose per layer from measured attention statistics, use a different operator for each regime, and accept a configuration only if the worst calibration dataset stays within a budget.

\paragraph{Layer roles in VGGT and our position.}
The closest work is RegimeVGGT~\cite{regimevggt}. It splits VGGT into three contiguous bands of layers (shallow layers without cross-view structure, middle layers that align views, deep layers needed for pose) and compresses them in a U-shape, with banded merging and key/value downsampling that protect a reference-frame anchor. Other studies probe layer roles~\cite{stary,vggtoverlap,bratulic,goodtoken,attncollapse,vggtslam2}; SAF3R~\cite{saf3r} finds single heads that attend to the anchor image. A layer-role view is therefore not new, and earlier methods already protect the first frame. We differ in four ways. Our regimes are measured per layer and are not contiguous. We show that anchoring is a whole-layer effect tied to the first position (52--62\% of the attention) and absent in $\pi^3$. We choose an operator per regime and a numeric format per layer under the calibration check. And we measure on a 12\,GB GPU, with the gains of the engine and the method reported separately.
\emph{Quantisation:} QuantVGGT~\cite{quantvggt}, QVGGT~\cite{qvggt}, FGQ~\cite{fgq} and VersaQ-3D~\cite{versaq3d} quantise VGGT's linear layers; LiteVGGT~\cite{litevggt} and Lite3R~\cite{lite3r} add FP8 after fine-tuning. We keep the weights and set the precision of attention per layer.

\paragraph{Efficient attention.}
FlashAttention~\cite{flashattn,flashattn2,flashattn3} computes exact attention with tiling and online softmax. SageAttention~\cite{sageattn,sageattn2,sageattn2pp,sageattn3} quantises $QK^\top$ to INT8 and $P{\cdot}V$ to FP8 or FP4; SpargeAttn~\cite{spargeattn} predicts block sparsity; FlexAttention~\cite{flexattention} compiles attention variants; FP8 formats follow~\cite{fp8formats}. ToMe~\cite{tome} introduced proportional attention ($+\log s$ for a key representing $s$ tokens). Attention sinks~\cite{streamingllm,registers} absorb attention in a few tokens; anchoring is a frame-level analogue. These methods were developed for language models and video generators, where most of a token's attention falls on a few keys. In multi-view 3D models it does not (\cref{sec:analysis}), and we design the approximation around that.

\section{Analysis: three regimes of global attention}
\label{sec:analysis}
This section studies where the global attention of multi-view transformers goes, and uses the result to explain why existing accelerators fail.
\subsection{Setting and statistics}
We call each input image a \emph{frame}. \vggto~\cite{vggtomega} splits each of the $N$ frames into $16{\times}16$-pixel patches and adds 17 \emph{special tokens} (one camera token and 16 register tokens, extra learned tokens that collect global information~\cite{registers}), giving $T{=}689$ tokens per frame at $512{\times}336$ pixels. Each of its 24 blocks has a \emph{frame-attention} layer, in which tokens attend only within their own frame, and a \emph{global} layer, in which every token attends to all $NT$ tokens of all frames. Five global layers (blocks 2, 6, 9, 14, 20) attend only among special tokens in the released model (\emph{register-only} layers) and are never changed. The other 19 cost $O(N^2T^2d)$; at 50 frames global attention takes 42.5\% of the GPU time of our CUDA engine with exact attention (the \emph{exact engine}, \cref{sec:engine}; \cref{tab:decomp}).

During one forward pass with exact attention, we record five statistics for 1024 random patch queries per block and head: the \emph{normalised entropy} $H/\log(\#\text{keys})$ (1: equal weight on every key; 0: all weight on one key); the \emph{mass on frame 0}, the share of attention that queries of other frames give to frame 0 ($1/N$ if frame identity did not matter); the \emph{top-1 frame share}, the share of attention that goes to the most-attended frame; the entropy of the per-frame shares; and the effective number of attended keys, $e^H$. We do this for \vggto, VGGT-1B~\cite{vggt} and $\pi^3$~\cite{pi3} (bf16, official weights) on seven datasets (ETH3D, CO3Dv2, RealEstate10K, NRGBD, TUM-dynamic, 7-Scenes, Mip-NeRF~360) at $N{=}10$, 50 and, where memory allows, 200.

\begin{figure*}[t]
\centering
\includegraphics[width=\linewidth]{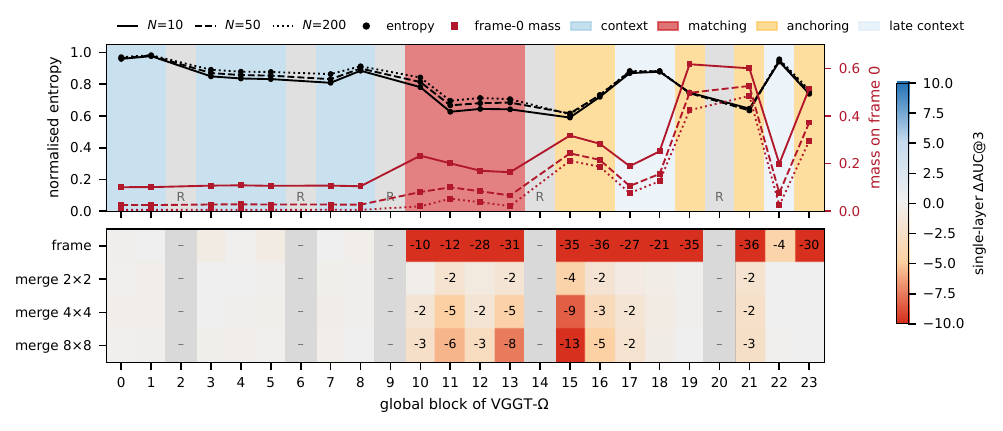}
\vspace{-1.8em}
\caption{\textbf{Regimes and tolerance in \vggto.} Top: normalised entropy (black) and attention mass on frame 0 (red) per global block at 10, 50 and 200 frames; background: regime at $N{=}10$. Bottom: $\Delta$AUC@3 when one block alone uses frame attention or merging (ETH3D and 7-Scenes); numbers shown for drops of at least 1.5. Grey ``R''/``--'' columns: register-only blocks, never changed.}
\label{fig:anatomy}
\end{figure*}

\subsection{Three regimes}
\Cref{fig:anatomy} (top) shows the statistics of \vggto. Every global layer looks at other frames about as much as it can (cross-frame mass 0.79--0.90 at 10 frames, where uniform is 0.90); what differs is \emph{how selectively}.
\begin{itemize}\setlength\itemsep{1pt}
\item \textbf{Diffuse context (blocks 0--8).} Entropy 0.81--0.98: each query spreads its attention almost evenly over all frames, and frame 0 gets only its uniform share (1.0--1.1\x). These layers compute scene-wide averages, not correspondences. We call them \emph{context layers}.
\item \textbf{Selective matching (blocks 10--16 except register-only 14; 15 and 16 also anchor).} Attention is concentrated (mean entropy 0.59--0.78; down to 0.25 for the most selective head), yet the most-attended frame receives only 0.26--0.48 of it: each query collects evidence from many frames. We call them \emph{matching layers}.
\item \textbf{Reference anchoring (blocks 19, 21, 23; partly 15, 16).} In these whole layers, not only single heads~\cite{saf3r}, 52--62\% of every query's attention goes to frame 0, 5--6\x its uniform share at 10 frames. Attention sinks in language models~\cite{streamingllm,registers} are single tokens; here the target is a whole frame, the one whose coordinates the outputs use. The effect is tied to the first position: if the middle frame is moved to the first slot, that slot still receives the same mass ($|\Delta|\le0.005$; supp.). We call them \emph{anchoring layers}.
\end{itemize}
Blocks 17, 18 and 22 are diffuse again; we call them \emph{late context layers}. They tolerate merging but not frame attention (\cref{fig:anatomy}, bottom), so frame attention is used only in the \emph{context stage}, the context layers before the first matching layer. We call this per-layer labelling the \emph{regime map}; it is the same in all seven datasets except three blocks near a threshold (7, 10, 16).

\paragraph{Attention is not sparse.} In no block of any of the three models does the heaviest frame hold more than 0.63 of the mass, and the effective number of attended keys is 9--83\% of all keys. Single heads can be peaky~\cite{saf3r}; whole blocks are not.

\paragraph{The regimes do not depend on sequence length.} From 10 to 200 frames, normalised entropy changes by at most 0.07 per block, and anchoring does not dilute: frame 0 still receives 42--48\% of the mass in blocks 19 and 21 at 200 frames. Only the matching layers spread their evidence over more frames (top-1 frame share 0.31 at 10 frames, 0.06 at 200).

\paragraph{The regimes differ between models.} All three models start with 9--11 context blocks and put their matching layers in the middle third (\cref{fig:teaser}a). \vggto then anchors strongly, VGGT-1B weakly (frame-0 mass 0.32--0.41), and $\pi^3$, which has no reference frame, not at all (supp.).

\subsection{Why sparsity, routing and merging fail}
\label{sec:why-fail}
The regimes account for every failure we observed when applying accelerators from language and video models to \vggto:
\begin{itemize}\setlength\itemsep{1pt}
\item \textbf{Block sparsity finds nothing to skip.} SparseVGGT~\cite{sparsevggt} reaches 7--8\% effective sparsity on \vggto at its default setting: it keeps accuracy but is only 1.17\x faster (\cref{sec:exp-rivals}).
\item \textbf{Routing to a few frames has little to select.} Routing lets each query attend only to the $k$ frames that seem most relevant. But in matching layers the frame that receives the most attention holds only 0.26--0.48 of a query's attention at 10 frames, and 0.06 at 200. Our port of BlazeVGGT-style routing (each frame attends to the half of the frames with the highest camera-token score, every global layer, 200 frames) loses 17 AUC@3 on the worst dataset (the \emph{worst-dataset $\Delta$}), and 37 with a quarter (supp.); BlazeVGGT~\cite{blazevggt} and Spark3R~\cite{spark3r} themselves target 1,000--2,000 frames.
\item \textbf{Merging or frame attention in matching layers destroys correspondence.} Frame attention on any single block in 10--23 costs 4--36 AUC@3 (\cref{fig:anatomy}), and merging up to 9 in blocks 10--16. Every rival that loses accuracy merges or drops keys there.
\item \textbf{Summaries are not enough.} Exchanging information across frames only through camera and register tokens costs 25 AUC@3 on ETH3D.
\item \textbf{The reference frame must keep all its keys.} Merging frame 0 together with the others in the single anchoring block 19 costs 2.7 AUC@3 on ETH3D, four times the cost of all \raf approximations together (\cref{tab:abl}). FastVGGT and OVGGT already protect first-frame tokens~\cite{fastvggt,ovggt}; we show why: the anchor is a whole-layer, positional regime.
\end{itemize}

\paragraph{Low-precision attention probabilities.} In context layers many attention probabilities are tiny, so one might expect them to round to zero in FP8. Few do, because the kernel scales each row by its maximum: entries that round to zero carry only 0.07\% of the mass, and the error of $P{\cdot}V$ (0.38\%) \emph{falls} as $N$ grows (supp.).

\section{\ra: regime-aware acceleration}
\label{sec:method}
\ra has three parts. Each layer gets an attention operator (\cref{sec:ops}) and a numeric format (\cref{sec:formats}); we call an assignment of operators to layers a \emph{schedule}. A calibration check tests the complete configuration and accepts it only if the worst calibration dataset stays within an accuracy budget (\cref{sec:verify}). Schedules come from the \emph{threshold schedule}, computed from one pass of exact attention (\cref{alg:rule}), or from a greedy search under the check. The regimes explain which operator each layer tolerates (\cref{tab:abl}); the calibration check, not the regimes, picks the named settings.

\subsection{Operator per regime}
\label{sec:ops}
All operators keep every query and leave the weights unchanged.
\textbf{Dense} (matching layers): attention over all $NT$ keys.
\textbf{Frame} (context layers of the context stage only): each frame attends only to itself, \ie tokens are reshaped from $[B, NT]$ to $[BN, T]$ as in AVGGT~\cite{avggt}; cost $O(NT^2d)$. When a layer only averages and the frames carry similar information, each frame's own average is a good substitute for the average over all frames. This holds in the context stage but not in the late context layers (\cref{fig:anatomy}).
\textbf{Merge$_4$} (context and anchoring layers): for frames $1..N{-}1$, keys and values are reduced to the 17 special tokens plus $4{\times}4$ mean-pooled patch keys and values (after QK-norm), so a non-reference frame contributes 65 keys instead of 689. Frame 0 stays complete. With $c_j$ tokens pooled into key $j$,
\begin{equation}
\mathrm{Attn}(q) = \sum_j \frac{\exp(q^\top k_j/\sqrt d + \log c_j)}{\sum_{j'}\exp(q^\top k_{j'}/\sqrt d + \log c_{j'})}\, v_j ,
\label{eq:merge}
\end{equation}
the proportional attention of ToMe~\cite{tome} and GPA-VGGT~\cite{gpavggt}. For a uniform attention row, \cref{eq:merge} is exact: the pooled key carries exactly the mass of the $c_j$ keys it replaces. Intuitively, if a query weights 16 neighbouring patches about equally, one averaged key that counts 16 times gives the same output. This is why merging suits context layers, and why keeping frame 0 complete makes it suitable for anchoring layers, whose selective mass sits on frame 0.

\subsection{Numeric format}
\label{sec:formats}
All attention uses INT8 $QK^\top$ with key smoothing and per-block scales, following SageAttention~\cite{sageattn,sageattn2}, and FP8 E4M3 $P{\cdot}V$ with FP16 accumulation in each 64-key block, folded into FP32 (SageAttention2++~\cite{sageattn2pp}). \Cref{sec:analysis} shows that this kernel's own error is small in every regime. A layer whose output is unusually sensitive downstream gets \textbf{16-bit $P{\cdot}V$}: FP16 $P$ and $V$ with FP16 accumulation per 64-key block, which runs at the full FP16 tensor-core rate on GeForce Ada. Per global layer at 50 frames, the two formats take 23.1 and 34.2\,ms, against 50.0\,ms for exact bf16 attention. Which layers need 16-bit is the one choice the statistics do not predict, so the calibration check decides it (\cref{sec:verify}). Merge and frame operators use the layer's format.

\subsection{Threshold schedule}
\label{sec:rule}
\begin{algorithm}[t]
\caption{Threshold schedule (from one pass of exact attention on 10-frame calibration sequences)}
\label{alg:rule}
\small
\begin{algorithmic}[1]
\Require per-block entropy $e_\ell$, minimum head entropy $e^{\min}_\ell$, frame-0 mass $f_\ell$ and its ratio to uniform $r_\ell$
\State \textbf{label} $\ell$: matching if $e_\ell < 0.80$; else anchoring if $f_\ell \ge 0.28$; else context
\State \textbf{context stage} $\gets$ blocks before the first one that is matching, anchoring, or has $r_\ell \ge 1.5$
\For{each global block $\ell$ (register-only blocks skipped)}
  \If{$\ell$ context and in the context stage} $o_\ell \gets$ merge$_4$ (or frame)
  \ElsIf{$\ell$ context and $e^{\min}_\ell \ge 0.75$} $o_\ell \gets$ merge$_4$
  \ElsIf{$\ell$ anchoring and $e^{\min}_\ell \ge 0.40$} $o_\ell \gets$ merge$_4$, frame 0 complete
  \Else\ $o_\ell \gets$ dense
  \EndIf
\EndFor
\end{algorithmic}
\end{algorithm}
\Cref{alg:rule} turns the regime map into a schedule. Its two tests follow the analysis: an operator that removes cross-frame keys is safe only if the cross-frame mass is spread evenly over the frames, and merging is safe only if \emph{every} head stays diffuse, because one selective head suffices to break it (\cref{sec:exp-rule}). The context stage ends at the first block whose frame-0 mass reaches 1.5\x its uniform share; this keeps block 10, which lies on the entropy threshold at long $N$, dense. We call the variant with merge$_4$ in the context stage the \emph{threshold schedule}; the variant with frame attention there is faster and less accurate (supp.). On \vggto \cref{alg:rule} returns the same schedule from 10- and 50-frame statistics: merge$_4$ on blocks 0, 1, 3, 4, 5, 7, 8, 19 and 22. It needs one dense pass of about 4\,s per calibration sequence.

\paragraph{How the thresholds were set.} We chose them once, after inspecting \vggto's single-layer sensitivity on the calibration datasets, so their agreement with that map (\cref{sec:exp-rule}) is a fit, not a test. Refits on calibration-only statistics move them substantially but return the same schedule (supp.); they do not transfer to other models (\cref{sec:exp-gen}).

\subsection{Calibration check and search}
\label{sec:verify}
\paragraph{Data.} Calibration uses five datasets: Mip-NeRF~360 (4 scenes), CO3Dv2 (banana, donut), RealEstate10K (30 sequences), 7-Scenes training and TUM-dynamic validation sequences. Evaluation (\cref{sec:exp}) is \emph{sequence-disjoint} from calibration everywhere, but \emph{dataset-disjoint} only for ETH3D and NRGBD; the supplement lists every sequence. A \emph{frame sample} is one draw of 10 frames per sequence.

\paragraph{The check.} Single-layer costs do not add up: $2{\times}2$ merging of block 12 costs nothing alone on outdoor calibration scenes ($+0.01$ AUC@3) but 1.05 when added to other approximations (supp.). A schedule must therefore be checked as a whole. We evaluate the complete configuration on the five calibration datasets and accept it only if the accuracy drop on the \emph{worst} of them stays within a budget ($-0.5$ AUC@3 and $-0.3$ AUC@30 when choosing numeric formats, $-1$ and $-0.5$ when choosing operators). We use the worst dataset rather than the mean, so that a loss on one dataset cannot hide behind gains on others; in an earlier protocol, calibrating on outdoor scenes alone cost 1.73 AUC@3 on the indoor RGB-D evaluation data (supp.).

\paragraph{Precision.} Starting from FP8 $P{\cdot}V$ everywhere, layers are given 16-bit $P{\cdot}V$ in order of their single-layer gain until the budget is met. This stops after two layers, block 15 (matching and anchoring) and block 4 (context), and gives \textbf{\ra}.

\paragraph{Schedule.} Independently of \cref{alg:rule}, a greedy search starts from \ra, visits layers by single-layer cost, tries frame attention before merging on each layer, and accepts an operator only if the cumulative configuration stays within the budget. The search itself is \emph{not} restricted to the regime map: every non-register layer and both operators are candidates. It needs 49 configurations (38 single-layer evaluations, 11 greedy trials) and accepts frame attention on blocks 0, 4, 5 and merge$_4$ on blocks 3 and 19, all of them context or anchoring layers. This schedule is \textbf{\raf}. It was fixed on one frame sample before any evaluation run; no evaluation data entered the choice. That the search avoids matching layers does not by itself confirm the regime map, because these layers are also the most sensitive ones. A greedy search that uses only single-layer sensitivity, with no regime statistics and no preference for frame attention, run after our evaluation (all 38 layer--operator pairs, same budget; supp.), also picks only context and anchoring layers (merge$_4$ on 0, 1, 3, 4, 8, 19, 22; frame on 5), close to the threshold schedule, but it fails at long sequences (\cref{sec:exp-main}).

\paragraph{Protocol sensitivity.} On the original and two new frame samples, \ra passes the operator budget on all three (the stricter precision budget on the original only), \raf on two, and the threshold schedule on none, so we report the latter as unverified. A stricter three-sample search finds a schedule that is more accurate at most, but not all, sequence lengths (supp.); no schedule dominates. \raf, fixed before evaluation, remains our named setting: the fastest schedule found at the $-1$ budget, not the most accurate one.

\subsection{Native engine}
\label{sec:engine}
PyTorch SDPA falls back to slower kernels when given the bias of \cref{eq:merge}, and FlexAttention~\cite{flexattention} cannot take INT8/FP8 operands, so we run the model in a native CUDA engine: cuBLASLt bf16 matrix multiplications, our own attention kernels (INT8/FP8, 16-bit, merged, frame), and a static memory planner. We call this engine with exact bf16 attention the \textbf{exact engine}. Comparing \ra and \raf with the exact engine measures the method alone; comparing the exact engine with Official \vggto measures the engine alone (\cref{tab:decomp}); their accuracy differs only on ETH3D (0.8 AUC@3, supp.). Kernel details are in the supplement.

\section{Experiments}
\label{sec:exp}
\paragraph{Setup.} We use the released \vggto checkpoint at 512\,px maximum side. Timings are on one RTX 4070 SUPER (12\,GB, CUDA 12.8) with no other process on the GPU: full forward (trunk, camera and dense heads), CUDA events, Mip-NeRF~360 kitchen at $512{\times}336$, median of 5--10 runs after 2 warm-ups. Official \vggto is the released PyTorch code (fp32 weights, bf16 autocast).
\emph{Accuracy} uses six evaluation datasets (\cref{sec:verify}): ETH3D (13 scenes, official frames), CO3Dv2 (40 sequences, 8 unseen categories), RealEstate10K (38 scenes), 7-Scenes, NRGBD and TUM-dynamic, 10 frames per sequence and three frame-sampling seeds. Pose accuracy follows \vggto's protocol: for every pair of frames we measure the angular error of the predicted relative rotation and of the relative translation direction. \textbf{AUC@3} is the area under the curve of the fraction of pairs whose larger error is below a threshold, for thresholds up to 3$^\circ$ (\textbf{AUC@30}: up to 30$^\circ$), in points from 0 to 100; higher is better. The \textbf{worst-dataset $\Delta$} is the lowest change among the six evaluation datasets. Differences below about 0.5 AUC@3 are seed noise; every $\pm$ is the sample standard deviation (ddof${=}1$) over the three seeds. Depth accuracy ($\delta_{1.25}$, the share of pixels within 25\% of the true depth, and AbsRel, the mean relative error) stays within 0.02 points and 0.0002 of Official for both settings (supp.).

\begin{table*}[t]
\centering\small
\setlength{\tabcolsep}{3.4pt}
\caption{\textbf{Operating points on \vggto.} $\Delta$AUC@3 against the exact engine on six evaluation datasets ($N{=}10$), mean over three frame-sampling seeds. Worst: per-seed minimum over the datasets, mean\,$\pm$\,std over seeds; vs Off.: worst $\Delta_3$ against Official \vggto (\cref{tab:rivals}). Latency: full forward, RTX 4070 SUPER, mean of timing sessions (supp.); \x official: against Official at 3605\,ms (out of memory at $N{\ge}100$). Cal.: calibration samples (of three) passing the check. The sensitivity-only greedy search is post hoc; at 100/200 frames it loses 3.42/2.10 on RE10K, against 1.20/1.03 for \raf (supp.).}
\label{tab:main}
\resizebox{\textwidth}{!}{\begin{tabular}{@{}lcccccc|cccc|cc|cc@{}}
\toprule
 & ETH3D & CO3D & RE10K & 7-Sc. & NRGBD & TUM-dyn & worst $\Delta_3$ & worst $\Delta_{30}$ & vs Off. & cal. & \multicolumn{2}{c|}{$N{=}50$} & \multicolumn{2}{c}{$N{=}200$} \\
 & & & & & & & & & & & ms & \x official & s & \x exact \\
\midrule
Exact engine (AUC@3) & 49.91 & 51.21 & 38.49 & 21.11 & 85.46 & 39.66 & 0 & 0 & $-0.86$ & -- & 2237 & 1.61 & 20.74 & 1.00 \\
\midrule
\textbf{\ra} (precision only) & $-$0.23 & $-$0.15 & $-$0.19 & $+$0.01 & $-$0.08 & $-$0.22 & $\mathbf{-0.51}${\scriptsize$\pm$0.11} & $\mathbf{-0.24}$ & $-1.03$ & 3/3 & 1734 & 2.08 & 12.51 & 1.66 \\
\textbf{\raf} (+ schedule) & $-$0.34 & $+$0.07 & $-$0.22 & $-$0.56 & $-$0.49 & $-$0.12 & $\mathbf{-0.80}${\scriptsize$\pm$0.17} & $-1.20$ & $-1.14$ & 2/3 & 1616 & 2.23 & 10.56 & 1.97 \\
\midrule
\quad threshold schedule & $+$0.17 & $-$0.42 & $+$0.24 & $-$0.22 & $-$0.36 & $-$0.84 & $-0.99${\scriptsize$\pm$0.54} & $-1.23$ & $-1.16$ & 0/3 & 1544 & 2.33 & 9.39 & 2.21 \\
\quad greedy search, budget $-2$ & $-$1.03 & $-$0.10 & $-$0.45 & $-$0.73 & $-$1.48 & $-$0.16 & $-1.59${\scriptsize$\pm$0.50} & $-0.75$ & $-1.87$ & -- & 1547 & 2.33 & 9.51 & 2.18 \\
\quad sensitivity-only greedy & $+$0.40 & $-$0.27 & $-$0.61 & $-$0.51 & $-$0.03 & $-$0.40 & $-0.91${\scriptsize$\pm$0.25} & $-0.21$ & -- & 2/3 & 1558 & 2.31 & 9.67 & 2.14 \\
\bottomrule
\end{tabular}}
\end{table*}
\begin{table}[t]
\centering\small
\setlength{\tabcolsep}{3pt}
\caption{\textbf{Engine vs method.} Kernel time by category (Nsight Systems, 99--100\% of the trunk; global attention grouped by block range: 0--9 and 22 are context, 10--13, 15, 16 matching (14 register-only), 17--21 and 23 anchoring and late context), RTX 4070 SUPER, kitchen $512{\times}336$. Components: one Nsight session (they sum to 2232, 1736, 1619\,ms); total: timed sessions as in \cref{tab:main}; step \x: over the previous column, Official \vggto 3605\,ms at $N{=}50$, OOM at 200.}
\label{tab:decomp}
\setlength{\tabcolsep}{2.2pt}
\resizebox{\linewidth}{!}{\begin{tabular}{@{}lccc|ccc@{}}
\toprule
 & \multicolumn{3}{c|}{$N{=}50$ (ms)} & \multicolumn{3}{c}{$N{=}200$ (s)} \\
component & Exact & \ra & Fast & Exact & \ra & Fast \\
\midrule
patch embedding (DINO) & 402 & 402 & 402 & 1.63 & 1.63 & 1.63 \\
frame attention & 28 & 28 & 28 & 0.11 & 0.11 & 0.11 \\
global attn., blocks 0--9, 22 & 400 & 197 & 100 & 6.61 & 3.17 & 1.55 \\
global attn., blocks 10--13, 15, 16 & 300 & 150 & 150 & 4.95 & 2.41 & 2.41 \\
global attn., blocks 17--21, 23 & 248 & 116 & 97 & 4.13 & 1.85 & 1.52 \\
GEMMs (QKV, proj., MLP) & 533 & 529 & 520 & 2.05 & 2.07 & 2.05 \\
norms, activations, gaps & 202 & 197 & 203 & 0.76 & 0.78 & 0.79 \\
heads, upload & 119 & 118 & 119 & 0.50 & 0.50 & 0.50 \\
\midrule
total & 2237 & 1734 & 1616 & 20.74 & 12.51 & 10.56 \\
step \x\ (engine; precision; schedule) & 1.61 & 1.29 & 1.07 & -- & 1.66 & 1.19 \\
\bottomrule
\end{tabular}}
\end{table}
\subsection{Do the regimes predict tolerance?}
\label{sec:exp-rule}
We score each statistic by its ranking accuracy: the probability that it ranks a tolerant layer (alone, at most 1 worst-dataset AUC@3 lost) above an intolerant one (supp.). An even spread of cross-frame mass predicts which layers survive frame attention (0.91 on \vggto); the least diffuse head predicts which survive merging (0.91 on \vggto, 0.97 on VGGT-1B). On the 38 layer--operator pairs, \cref{alg:rule} never keeps a tolerant pair dense and approves 3 intolerant ones, each at most 0.7 points beyond the budget; this is in-sample on \vggto.

\begin{table}[t]
\centering\small
\setlength{\tabcolsep}{2.6pt}
\caption{\textbf{Numeric format} of all 19 dense global layers, $\Delta$ against the exact engine over six evaluation datasets (seed 0); latency at $N{=}50$, one timing session. All rows use INT8 $QK^\top$ except the first two.}
\label{tab:prec}
\resizebox{\linewidth}{!}{\begin{tabular}{@{}lcccr@{}}
\toprule
$P{\cdot}V$ format (accumulation) & worst $\Delta_3$ & worst $\Delta_{30}$ & mean $\Delta_3$ & ms \\
\midrule
bf16 (exact engine) & 0 & 0 & 0 & 2237 \\
FP8 (FP32), SageAttention2 & $-$2.14 & $-$0.85 & $-$0.91 & 1799 \\
FP8 (FP16), SageAttention2++ & $-$0.56 & $-$1.12 & $-$0.38 & 1713 \\
FP8 (FP32), row sum of rounded $P$ & $-$0.41 & $-$0.12 & $+$0.01 & 1811 \\
16-bit (FP16) & $-$0.21 & $-$0.06 & $-$0.01 & 1943 \\
\midrule
\textbf{\ra}: FP8 (FP16), 16-bit on 4, 15 & $-$0.41 & $-$0.12 & $-$0.15 & 1738 \\
\bottomrule
\end{tabular}}
\end{table}
\begin{table}[t]
\centering\small
\setlength{\tabcolsep}{2.2pt}
\caption{\textbf{Accelerators on \vggto, same RTX 4070 SUPER}, one timing session (exact engine, \ra: three, $2237\pm2$, $1734\pm1$\,ms). Worst $\Delta$AUC against Official over six datasets as in \cref{tab:main}, or seed 0 ($^\ast$). $^\dagger$Released code ported onto \vggto, poses only; $^\ddagger$in our engine at \ra's precision. $\pm$0.00: ETH3D (one frame set) worst on every seed.}
\label{tab:rivals}
\setlength{\tabcolsep}{2.0pt}
\resizebox{\linewidth}{!}{\begin{tabular}{@{}lccccc@{}}
\toprule
 & s ($N{=}50$) & \x & worst $\Delta_3$ & worst $\Delta_{30}$ & s ($N{=}200$) \\
\midrule
Official \vggto & 3.61 & 1.00 & 0 & 0 & OOM \\
SparseVGGT$^\dagger$~\cite{sparsevggt}, default & 3.08 & 1.17 & $-$1.04{\scriptsize$\pm$0.26} & $-$0.51 & OOM \\
\quad CDF 0.90 & 3.06 & 1.18 & $-$0.74$^\ast$ & $-$0.64 & OOM \\
GPA-VGGT$^\dagger$~\cite{gpavggt}, default & 2.77 & 1.30 & $-$3.14{\scriptsize$\pm$1.08} & $-$0.89 & OOM \\
\quad $r{=}0.5$ on 9--23 & 2.85 & 1.27 & $-$4.65$^\ast$ & $-$1.92 & OOM \\
FastVGGT$^\dagger$~\cite{fastvggt} & 2.81 & 1.28 & $-$21.19$^\ast$ & $-$9.17 & OOM \\
SageAttention2$^\ddagger$~\cite{sageattn2}, uniform & 1.80 & 2.00 & $-$2.47$^\ast$ & $-$1.10 & 13.57 \\
frame 0--8, merge$_2$ 10--23$^\ddagger$ & 1.36 & 2.65 & $-$37.89{\scriptsize$\pm$0.00} & $-$28.41 & 6.63 \\
\quad frame 0--8 only & 1.56 & 2.31 & $-$2.97{\scriptsize$\pm$0.31} & $-$0.82 & 9.72 \\
GPA-style merge$^\ddagger$, \raf's layers & 1.65 & 2.18 & $-$1.03{\scriptsize$\pm$0.00} & $-$1.25 & 11.13 \\
\midrule
Exact engine & 2.24 & 1.61 & $-$0.86{\scriptsize$\pm$0.11} & $-$0.30 & 20.74 \\
\textbf{\ra} & 1.73 & 2.08 & $-$1.03{\scriptsize$\pm$0.00} & $-$0.39 & 12.51 \\
\textbf{\raf} & 1.62 & 2.23 & $-$1.14{\scriptsize$\pm$0.00} & $-$1.13 & 10.56 \\
\bottomrule
\end{tabular}}
\end{table}
\subsection{How much faster, and at what cost in accuracy?}
\label{sec:exp-main}
\Cref{tab:main} lists our operating points and \cref{tab:decomp} separates engine and method. \textbf{The method alone} (against the exact engine) is 1.29\x faster at 50 frames with low precision, and a further 1.07\x with the \raf schedule (1.38\x in total); at 200 frames these become 1.66\x and 1.19\x (1.97\x in total), because global attention then takes 76\% of the exact engine's time. \textbf{\ra}, which uses no schedule, stays within $0.51\pm0.11$ AUC@3 and 0.24 AUC@30 of the exact engine on the worst dataset, and \textbf{\raf} within $0.80\pm0.17$. Against Official \vggto, \ra is 2.08\x and \raf 2.23\x faster, at $-1.03$ and $-1.14$ worst-dataset AUC@3, 0.86 of which already comes from the exact engine (\cref{tab:rivals}). \raf and the threshold schedule lose 1.0--1.2 AUC@30 on ETH3D.
\emph{More frames.} With 50, 100 and 200 frames per sequence (five datasets; ETH3D is too short; three seeds), \ra stays within 0.32 and \raf within 1.20 AUC@3 of the exact engine on the worst dataset (supp.). On ETH3D at 50 frames, \raf loses 1.65 AUC@3 against the exact engine (2.22 against Official), and \ra nothing.

\begin{table}[t]
\centering\small
\setlength{\tabcolsep}{2.6pt}
\caption{\textbf{Ablations of \raf} at \ra's precision: worst $\Delta$AUC@3 against \ra over six datasets, mean\,$\pm$\,std of the per-seed worst over three seeds; mean $\Delta_3$: six-dataset mean. Rows d--d$_2$: five random draws (range of per-draw worst values). Rows d$_1$, d$_2$ are against the exact engine (\raf: $-0.80\pm0.17$): d$_1$ draws layers outside 10--16; d$_2$ draws frame layers from context blocks 0--8 and merge layers from context or anchoring blocks. Latency $N{=}50$, one timing session.}
\label{tab:abl}
\resizebox{\linewidth}{!}{\begin{tabular}{@{}lccr@{}}
\toprule
schedule & worst $\Delta_3$ & mean $\Delta_3$ & ms \\
\midrule
\ra (no schedule) & 0 & 0 & 1732 \\
\textbf{\raf}: frame 0, 4, 5; merge$_4$ 3, 19 & $-0.70${\scriptsize$\pm$0.19} & $-0.14$ & 1608 \\
\midrule
(a) merge $2{\times}2$ instead of $4{\times}4$ & $-0.80${\scriptsize$\pm$0.09} & $-0.12$ & 1620 \\
(b) no $+\log c$ bias & $-1.14${\scriptsize$\pm$0.17} & $-0.05$ & 1604 \\
(c) frame 0 merged in block 19 & $-2.68${\scriptsize$\pm$0.00} & $-0.73$ & 1606 \\
(d) same operators, random layers (5 draws) & $-34.6$ to $-85.4$ & $-39.1$ & 1621 \\
(e) same operators on matching blocks & $-85.08${\scriptsize$\pm$2.97} & $-46.74$ & 1609 \\
(e$'$) only the two merges moved to 12, 13 & $-19.60${\scriptsize$\pm$0.00} & $-8.88$ & 1613 \\
\midrule
\multicolumn{4}{@{}l}{\emph{against the exact engine}} \\
(d$_1$) random, non-matching layers & $-59.4$ to $-83.2$ & $-40.9$ & 1621 \\
(d$_2$) random, operator matched to regime & $-1.45$ to $-4.23$ & $-1.27$ & 1615 \\
\bottomrule
\end{tabular}}
\end{table}
\paragraph{Ablations.} \Cref{tab:abl} varies one choice at a time. Merging $2{\times}2$ instead of $4{\times}4$ changes nothing, the $+\log c$ bias gains 0.44 (edge of noise), and keeping frame 0 complete matters most. The same operators at random layers lose 35--85 AUC@3, and still 59--83 when drawn only outside the matching layers (d$_1$), because frame attention lands on a late layer; drawing each operator from its regime (d$_2$) loses only 1.5--4.2. The post-hoc sensitivity-only greedy search picks the same kind of layers and ties \raf at 10 and 50 frames, but loses 3.4/2.1 AUC@3 on RE10K at 100/200 frames (\raf: 1.2/1.0). Both pass two of three calibration samples, so neither the ranking nor the check at 10 frames predicts this long-sequence failure.

\subsection{Where does low-precision attention lose accuracy?}
\label{sec:prec}
\Cref{tab:prec} changes only the format of attention. Uniform SageAttention2 loses 2.14 AUC@3 on NRGBD; 16-bit $P{\cdot}V$ fixes it, finer $QK^\top$ quantisation does not. SageAttention2++ recovers most of it but loses 1.12 AUC@30 on TUM-dynamic. \ra, with 16-bit $P{\cdot}V$ on the two layers the check selects, is within 0.2 AUC@3 of uniform 16-bit at 0.89\x its latency. FP8 matrix multiplications in the linear layers cost 0.4--0.9 AUC@3 on every calibration dataset and are rejected by the check.

\subsection{How do published accelerators compare on the same GPU?}
\label{sec:exp-rivals}
\Cref{tab:rivals} and \cref{fig:teaser}b,c compare published accelerators on the same GPU and frames. SparseVGGT skips little: it matches \ra's worst-dataset accuracy ($-1.04$ vs $-1.03$) at 1.77\x the latency. GPA-VGGT and FastVGGT merge or drop keys in the matching layers and lose 3.1 and 21.2 AUC@3; \raf is 1.71\x faster than GPA-VGGT and loses 2.0 points less, partly through our engine. In our engine, GPA-style $2{\times}2$ merging on \raf's layers ties \raf, and the AVGGT/GPA-VGGT layer split (frame attention 0--8, merging 10--23) loses 37.9 (supp.).

\subsection{How many frames fit in 12\,GB?}
\label{sec:exp-scale}
Official \vggto runs at most 90 frames (8.4\,s) and runs out of memory at 100. \ra and \raf run 200 frames in 12.5\,s and 10.6\,s at a 9.7\,GiB peak (supp.). At 200 frames, the single-pass exact engine scores 32.8 AUC@3 on 7-Scenes against 28.4 for chunked \vggto (supp.); the memory gain is the engine's.
\subsection{Do the regimes and the method transfer to other models?}
\label{sec:exp-gen}
The regimes carry over to VGGT-1B and $\pi^3$; the thresholds of \cref{alg:rule} do not. In PyTorch, low-precision attention alone costs 0.84 (VGGT-1B) and 0.74 ($\pi^3$) worst-dataset AUC@3 at 1.36\x and 1.31\x speed-up (50 frames). The threshold schedule with \vggto's thresholds loses 7.8 (VGGT-1B) and 8.8 ($\pi^3$). A greedy search over the map's context and anchoring layers loses 3.79 at 1.53\x on VGGT-1B and 1.09 at 1.39\x on $\pi^3$ (supp.).

\section{Limitations and conclusion}
\label{sec:limits}
\textbf{Calibrated, not universal.} Each model needs its own search (\cref{sec:exp-gen}). On \vggto, \raf loses 2.2 AUC@3 on ETH3D at 50 frames, and no schedule is best at every length (\cref{sec:verify}).
\textbf{Coverage.} Only ETH3D and NRGBD are unseen in calibration; the other four share datasets, not sequences. Routing restricted to context or anchoring layers is untested.
\textbf{Rivals and hardware.} Published accelerators run as PyTorch ports; on a 96\,GB Blackwell GPU, GPA-VGGT outperforms \ra at 500 frames (supp.).

\noindent\textbf{Conclusion.} Multi-view global attention is structured, not sparse: layers average over the scene, match points across images, or anchor on the first image. Sensitivity finds the layers to approximate; the regimes explain which operator each tolerates and why frame 0 must stay complete. At 50 images \ra is 2.08\x faster than the released code, mostly via our engine; the method's gain grows with sequence length.\label{sec:endmain}

\section*{LLM Usage Statement}
Large language models were used extensively in this work, under the supervision of the authors. We used Anthropic's Claude models (as coding agents) for: \textbf{research and engineering}: writing most of the native engine and CUDA kernel code, the evaluation and benchmarking scripts, and the ports of the rival methods, and running and logging the experiments on our GPUs; \textbf{literature}: summarising related papers and checking citations against the cited works; and \textbf{writing}: drafting substantial parts of the text, tables and figure scripts of this paper and its supplement, and editing for language. The authors chose the research direction and the method, set the experimental protocols, checked the code, results and claims, and revised the text. Every number in the paper comes from a logged run in our results files. The authors take full responsibility for the content.

{\small
\bibliographystyle{ieeenat_fullname}
\bibliography{refs}
}

\clearpage
\appendix
\section*{Supplementary material}
This appendix is the supplementary material (``supp.'' in the main text); ``the paper'' below refers to Sections 1--6.

\paragraph{Names and metrics used here.} \textbf{\ra} (Regime-AWare ATTention Approximation) is the method of the main paper: low-precision attention whose per-layer format is set by the calibration check; \textbf{\raf} adds the searched operator schedule. The \textbf{exact engine} is our CUDA engine with exact bf16 attention; \textbf{Official \vggto} is the released PyTorch code. The \textbf{threshold schedule} is the output of Alg.~1 of the main paper. Context, matching and anchoring layers are the three regimes (main paper, Sec.~3). \textbf{AUC@3} (\textbf{AUC@30}) is the area under the curve of the fraction of frame pairs whose larger relative rotation or translation-direction error is below a threshold, for thresholds up to 3$^\circ$ (30$^\circ$), in points from 0 to 100. The \textbf{worst-dataset $\Delta$} is the lowest change among the evaluation (or calibration) datasets. $\delta_{1.25}$ is the share of pixels whose depth is within 25\% of the ground truth; AbsRel is the mean relative depth error.

\section{Extended related work}
\label{sec:s-rw}
\paragraph{Layer roles (prior work).} AVGGT~\cite{avggt} found that early global layers of VGGT do not match and converts them to frame attention, keeping frame 0 uncompressed when subsampling keys. RegimeVGGT~\cite{regimevggt} (NeurIPS 2026, first posted June 2026) finds three contiguous bands in VGGT (shallow layers without cross-view structure, middle layers that align views, deep layers that matter for pose) and compresses them in a U-shape with banded merging and protected K/V downsampling around a reference-frame anchor. Our deltas: (i) a per-layer, non-contiguous map on \vggto, VGGT-1B and $\pi^3$ rather than bands on VGGT; (ii) reference anchoring quantified as a whole-layer, positional regime that is absent in $\pi^3$; (iii) a different operator per regime, with the context stage alone tolerating frame attention; (iv) per-layer quantised attention set by a worst-dataset check; (v) same-GPU accounting of engine and method on a 12\,GB card. Our own controls (\cref{sec:s-ctrl}) show that which layers to approximate is also found by single-layer sensitivity alone. Probing studies analyse layer roles in DUSt3R-style models~\cite{stary} and co-visibility in VGGT, located in late layers~\cite{vggtoverlap}; Good Token Hunting~\cite{goodtoken} sets per-layer budgets from attention entropy; knockouts~\cite{bratulic} locate correspondence in middle layers.
\paragraph{Reference tokens.} FastVGGT~\cite{fastvggt} (ICLR 2026) keeps first-frame tokens as merge destinations, and OVGGT~\cite{ovggt} protects anchor tokens from cache eviction, consistent with our finding that frame 0 must stay exact in anchoring layers.
\paragraph{Long sequences and memory.} Chunked reconstruction~\cite{vggtlong,swiftvggt}, subscene decomposition~\cite{svggt} and streaming with a bounded KV cache by pruning, eviction, saliency-based admission or incremental fusion~\cite{streamvggt,infinitevggt,evict3r,ovggt,regvggt,incvggt} reduce memory for long inputs; we accelerate one full-attention pass and are complementary. Trained sparse attention~\cite{turbovggt} and concurrent routing methods~\cite{blazevggt,spark3r,vggtprime} target 1,000+ frames; our generic top-$k$ routing port (\cref{sec:s-ctrl}) is not their method.
\paragraph{Block sparsity.} SparseVGGT~\cite{sparsevggt} (CVPR 2026) replaces dense global attention by block-sparse attention; on \vggto at its default it finds 7--8\% effective sparsity (Sec.~3.3 of the paper).

\section{Configuration names}
\Cref{tab:s-names} maps the names used in the paper to the configuration strings of our engine, for reproducibility. A configuration string reads \texttt{<schedule>@<precision>\textasciitilde<per-layer overrides>}; \texttt{sagepp} is INT8~$QK^\top$ with FP8 $PV$ and FP16 accumulation, \texttt{sage} the same with FP32 two-level accumulation, \texttt{sage.pvh16} INT8~$QK^\top$ with 16-bit $PV$ and FP16 accumulation, \texttt{+fp8} FP8 GEMMs on blocks 16--23.
\begin{table*}[h]
\centering\small
\caption{Paper name $\to$ engine configuration.}
\label{tab:s-names}
\resizebox{\linewidth}{!}{\begin{tabular}{@{}ll@{}}
\toprule
paper name & configuration \\
\midrule
Official \vggto & official \texttt{eval/model.py} Predictor (fp32 weights, bf16 autocast) \\
Exact engine & \texttt{dense@bf16} \\
\ra & \texttt{dense@sagepp\textasciitilde L4=sage.pvh16/L15=sage.pvh16} \\
\raf (greedy search, one calibration sample, budget $-1$) & \texttt{L0:frame/L3:merge4/L4:frame/L5:frame/L19:merge4@sagepp\textasciitilde L15=sage.pvh16} \\
threshold schedule & \texttt{L0:merge4/L1:merge4/L3:merge4/L4:merge4/L5:merge4/L7:merge4/L8:merge4/L19:merge4/L22:merge4@sagepp\textasciitilde L4=sage.pvh16/L15=sage.pvh16} \\
3-sample search, first run (\cref{sec:s-sens}) & \texttt{L0:frame/L1:merge4/L4:merge4/L5:frame/L19:merge4@sagepp\textasciitilde L4=sage.pvh16/L15=sage.pvh16} \\
3-sample search, seed-matched (\cref{sec:s-sens}) & \texttt{L0:merge4/L3:merge4/L4:merge4/L5:frame/L8:merge4@sagepp\textasciitilde L4=sage.pvh16/L15=sage.pvh16} \\
threshold schedule, frame variant & \texttt{L0:frame/L1:frame/L3:frame/L4:frame/L5:frame/L7:frame/L8:frame/L19:merge4/L22:merge4@sagepp\textasciitilde L15=sage.pvh16} \\
greedy search, budget $-2$ (extra Pareto point) & \texttt{L0:frame/L1:merge4/L3:frame/L4:frame/L5:frame/L7:merge4/L8:merge4/L19:merge4@sagepp\textasciitilde L15=sage.pvh16} \\
earlier schedule point & \texttt{L0:merge4/L1:merge4/L3:merge4/L4:frame/L5:frame/L8:merge4@sagepp\textasciitilde L15=sage.pvh16} \\
FP8, rounded-$P$ row sum & \texttt{dense@sage.pq} \\
uniform SageAttention2 & \texttt{dense@sage} \\
uniform SageAttention2++ & \texttt{dense@sagepp} \\
uniform 16-bit $PV$ & \texttt{dense@sage.pvh16} \\
schedule only (v1, outdoor calibration) & \texttt{L0:merge4/L1:merge4/L3:frame/L4:frame/L5:frame/L7:merge4/L19:merge4} (\texttt{@sage+fp8}) \\
schedule only (multi-dataset) & \texttt{L0:merge4/L1:merge4/L3:merge4/L4:frame/L5:frame/L8:merge4} (\texttt{@sage+fp8}) \\
earlier fast point used in NVS & \texttt{L0:frame/L1:merge4/L3:merge4/L4:frame/L5:frame/L7:merge4/L8:merge4/L19:merge4@sage+fp8} \\
sensitivity-only greedy (\cref{sec:s-ctrl}) & merge$_4$ on 0, 1, 3, 4, 8, 19, 22, frame on 5, precision of \ra \\
AVGGT/GPA-style layer split (bf16) & frame on 0--8 (non-register), merge2 on 10--23 \\
AVGGT, published layers (in-engine control) & frame on 0--8, merge$_2$ $+\log c$ on 10--23 (non-register), precision of \ra \\
AVGGT, frame half only & frame on 0--8 (non-register), rest dense, precision of \ra \\
GPA-style merge on \raf's layers & merge$_2$ $+\log c$ on 0, 3, 4, 5, 19, precision of \ra \\
ablation modifiers (Tab.~5 of the paper) & \texttt{.nb}: no $+\log c$ bias on a merged layer; \texttt{.f0}: frame 0 pooled too \\
\bottomrule
\end{tabular}}
\end{table*}

\section{Regime maps of all three models}
\begin{figure*}[t]
\centering
\includegraphics[width=0.9\linewidth]{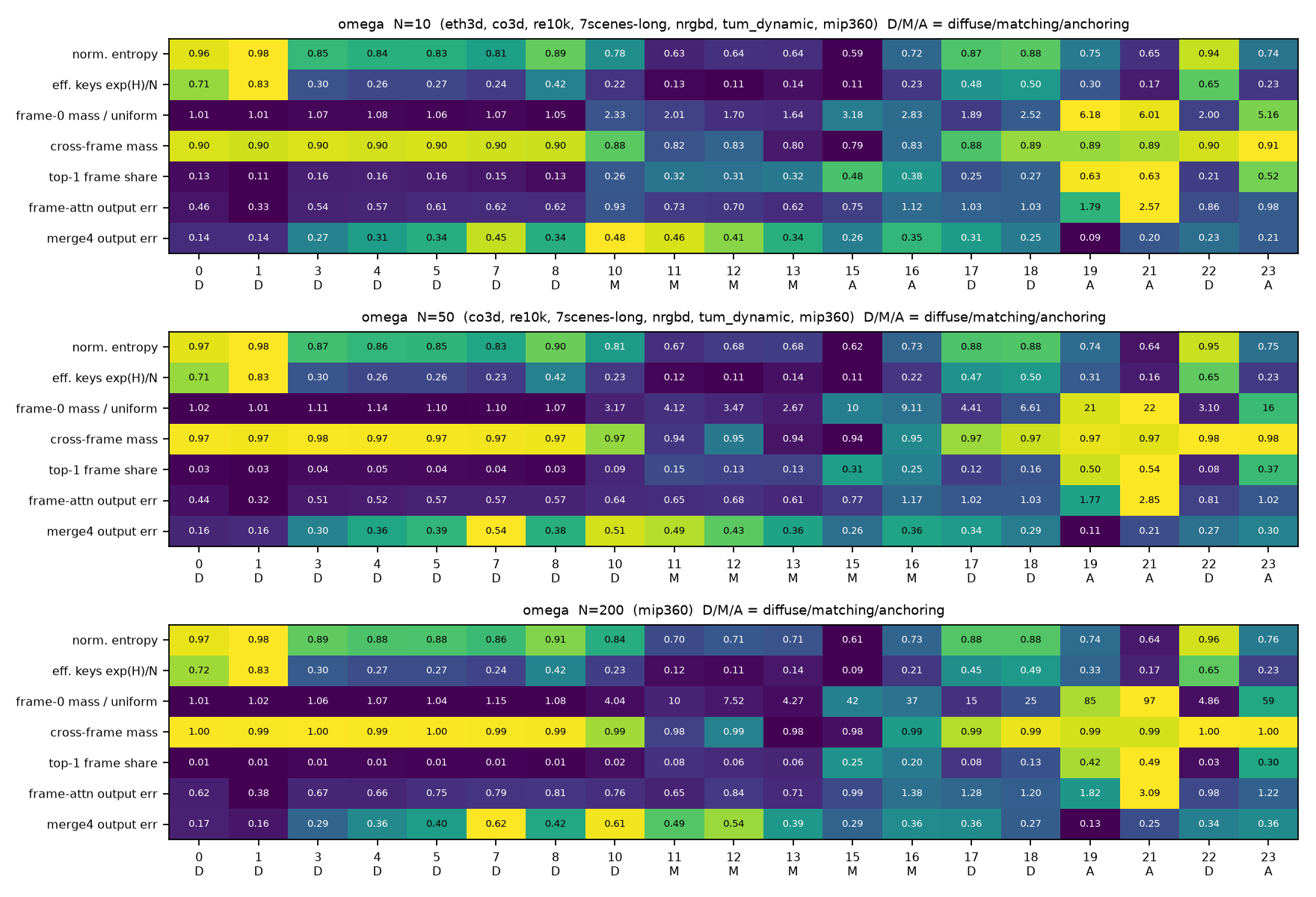}\\
\includegraphics[width=0.9\linewidth]{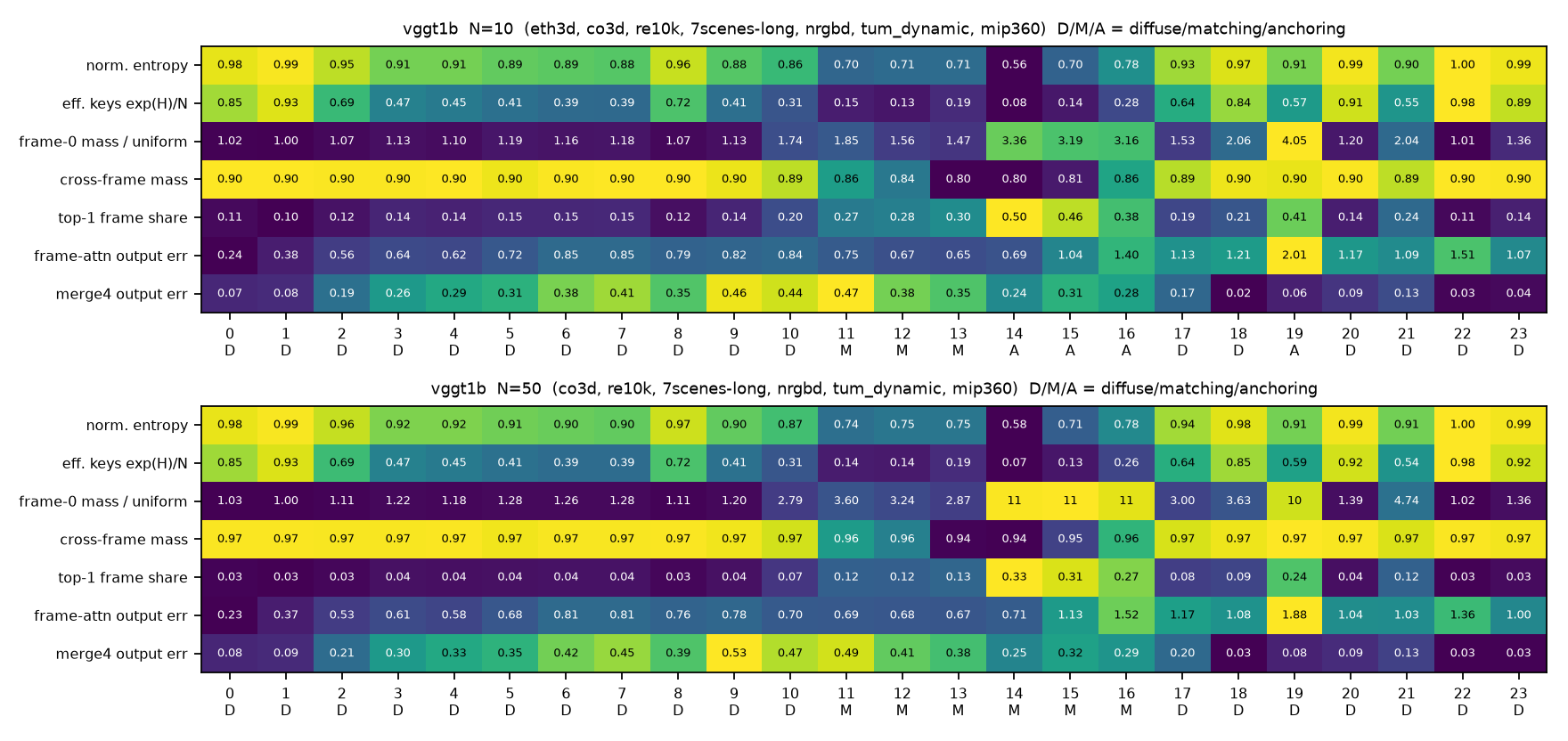}\\
\includegraphics[width=0.9\linewidth]{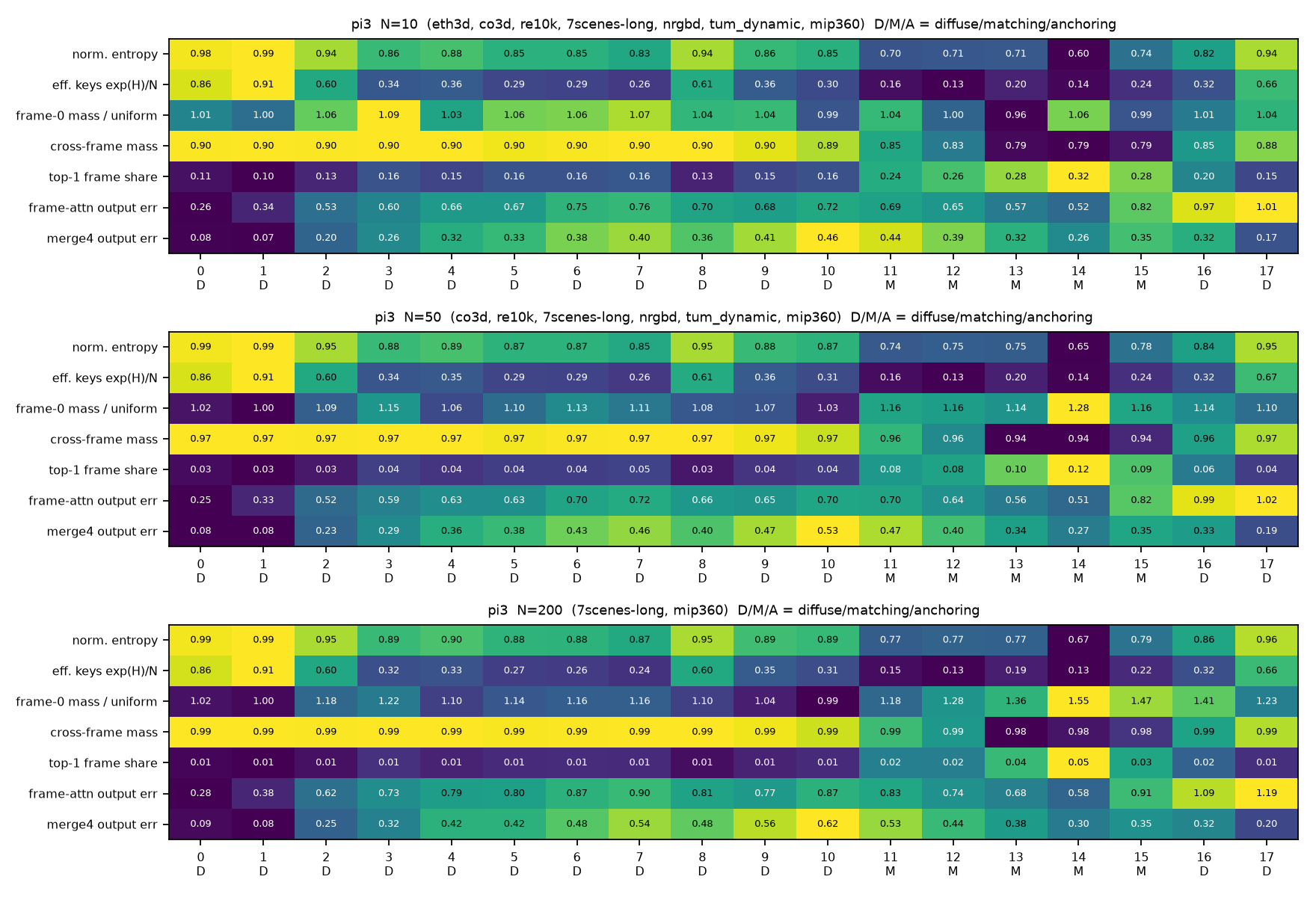}
\caption{Per-block attention statistics of \vggto (top), VGGT-1B (middle) and $\pi^3$ (bottom), one panel per sequence length.}
\label{fig:s-regimes}
\end{figure*}

\Cref{fig:s-regimes} shows the per-block statistics of \vggto, VGGT-1B and $\pi^3$ at each sequence length; the letter under each block is its label. \vggto at $N{=}200$ is measured on Mip-NeRF~360 only (7-Scenes at 200 frames and VGGT-1B at 200 frames run out of memory in the PyTorch statistics hook; the native engine runs these lengths). The thresholds of Alg.~1 are $t_\text{ent}{=}0.80$, $t_\text{anchor}{=}0.28$, $m_\text{diffuse}{=}0.75$, $m_\text{anchor}{=}0.40$. From the $N{=}200$ statistics, block 19 drops out of \vggto's schedule ($e^{\min}$ 0.40, on the threshold). On VGGT-1B, the anchoring threshold sits just above the $N{=}50$ frame-0 mass of blocks 14--16 and 19 (0.24--0.26); a ratio test ($\ge$10\x uniform at $N{=}50$) would label them consistently.
\begin{table}[t]
\centering\small
\setlength{\tabcolsep}{3pt}
\caption{\textbf{Statistics of one dense pass predict which layers tolerate an operator.} Rank AUC: probability that the statistic places a tolerant layer (single-layer worst-dataset $\Delta$AUC@3 $\ge -1$) on the tolerant side of an intolerant one; 1 is a perfect ranking. Statistics at $N{=}10$; at 50 and 200 frames they differ by $\le$0.03.}
\label{tab:predict}
\begin{tabular}{@{}lcccc@{}}
\toprule
 & \multicolumn{2}{c}{\vggto} & \multicolumn{2}{c}{VGGT-1B} \\
\cmidrule(lr){2-3}\cmidrule(lr){4-5}
statistic (tolerant direction) & frame & merge$_4$ & frame & merge$_4$ \\
\midrule
per-frame entropy (high) & \textbf{0.91} & 0.85 & \textbf{0.84} & 0.95 \\
frame-0 mass (low) & \textbf{0.91} & 0.80 & 0.81 & 0.85 \\
min.\ head entropy (high) & 0.80 & \textbf{0.91} & 0.78 & \textbf{0.97} \\
operator's own output error (low) & 0.87 & 0.72 & 0.70 & 0.67 \\
\bottomrule
\end{tabular}
\end{table}
\paragraph{Statistics vs tolerance.} \Cref{tab:predict} gives the rank AUC of each statistic against single-layer tolerance (Sec.~5.1 of the paper).
\paragraph{Threshold schedule vs single-layer map.} On \vggto (38 layer-operator pairs) the threshold schedule gives 13 TP, 3 FP (frame on blocks 1 and 3: $-1.11$/$-1.67$; merge$_4$ on 5: $-1.04$), 0 FN and 22 TN. On VGGT-1B the frame variant gives 21 TP, 5 FP, 2 FN, 20 TN. These counts are agreement with the map the thresholds were fitted on, not a test.

\paragraph{Threshold sensitivity and calibration-only refit.} \Cref{tab:s-thr} perturbs one threshold at a time on the published statistics ($N{=}10$, mean over seven datasets) and evaluates every new schedule on the six evaluation datasets (three seeds, precision of \ra). Refit: an exhaustive search over mid-points between observed statistic values, on the statistics of the calibration datasets only (Mip-NeRF~360, CO3Dv2, RE10K, TUM-dynamic; anatomy sequences, not the calibration sequences themselves) and the single-layer labels of the five calibration datasets, maximising correctly labelled pairs. It gives $t_\text{ent}{=}0.684$, $t_\text{anchor}{=}0.410$, $m_\text{diffuse}{=}0.737$, $m_\text{anchor}{=}0.421$ and exactly threshold schedule. Leave-one-calibration-dataset-out refits return the same schedule for object, indoor, RGB-D and dynamic; leaving out outdoor (Mip-NeRF~360) admits merge$_4$ on 17 and 23 ($-4.82\pm0.32$ held out, 1498\,ms). A zero-false-positive objective is too strict: it keeps merge$_4$ only on 0, 1, 3, 22 ($-0.65\pm0.48$, 1655\,ms, barely faster than \ra at 1732\,ms).
\begin{table}[t]
\centering\small\setlength{\tabcolsep}{2.2pt}
\caption{Threshold sensitivity of threshold schedule. Worst-dataset $\Delta$AUC@3 against \ra (six evaluation datasets, three seeds); latency $N{=}50$, one locked session. Perturbations not listed ($t_\text{ent}$ $-$10/$-$20\%, $t_\text{anchor}$ $-$10, $+$10, $+$20\%, $m_\text{diffuse}$ $\pm$10\%, $m_\text{anchor}$ $+$10\%) leave the schedule unchanged.}
\label{tab:s-thr}
\resizebox{\linewidth}{!}{\begin{tabular}{@{}llcr@{}}
\toprule
perturbation & change of merged blocks & worst $\Delta_3$ & ms \\
\midrule
none (threshold schedule) & 0, 1, 3, 4, 5, 7, 8, 19, 22 & $-0.78${\scriptsize$\pm$0.27} & 1538 \\
$t_\text{ent}$ $+$10\% & $-$3, 4, 5, 7, 8 & $-0.82${\scriptsize$\pm$0.29} & 1656 \\
$t_\text{ent}$ $+$20\% & $-$3, 4, 5, 7, 8, 22 & $-0.70${\scriptsize$\pm$0.14} & 1679 \\
$t_\text{anchor}$ $-$20\% & $+$10, 18 & $-8.59${\scriptsize$\pm$1.78} & 1502 \\
$m_\text{diffuse}$ $-$20\% & $+$17 & $-3.33${\scriptsize$\pm$0.56} & 1518 \\
$m_\text{diffuse}$ $+$20\% & $-$22 & $-0.73${\scriptsize$\pm$0.14} & 1559 \\
$m_\text{anchor}$ $-$10\%, $-$20\% & $+$16, 23 & $-10.03${\scriptsize$\pm$0.93} & 1501 \\
$m_\text{anchor}$ $+$20\% & $-$19 & $-0.72${\scriptsize$\pm$0.27} & 1561 \\
\bottomrule
\end{tabular}}
\end{table}

\section{FP8 probabilities in multi-view attention}
\begin{figure*}[t]
\centering
\includegraphics[height=4.2cm]{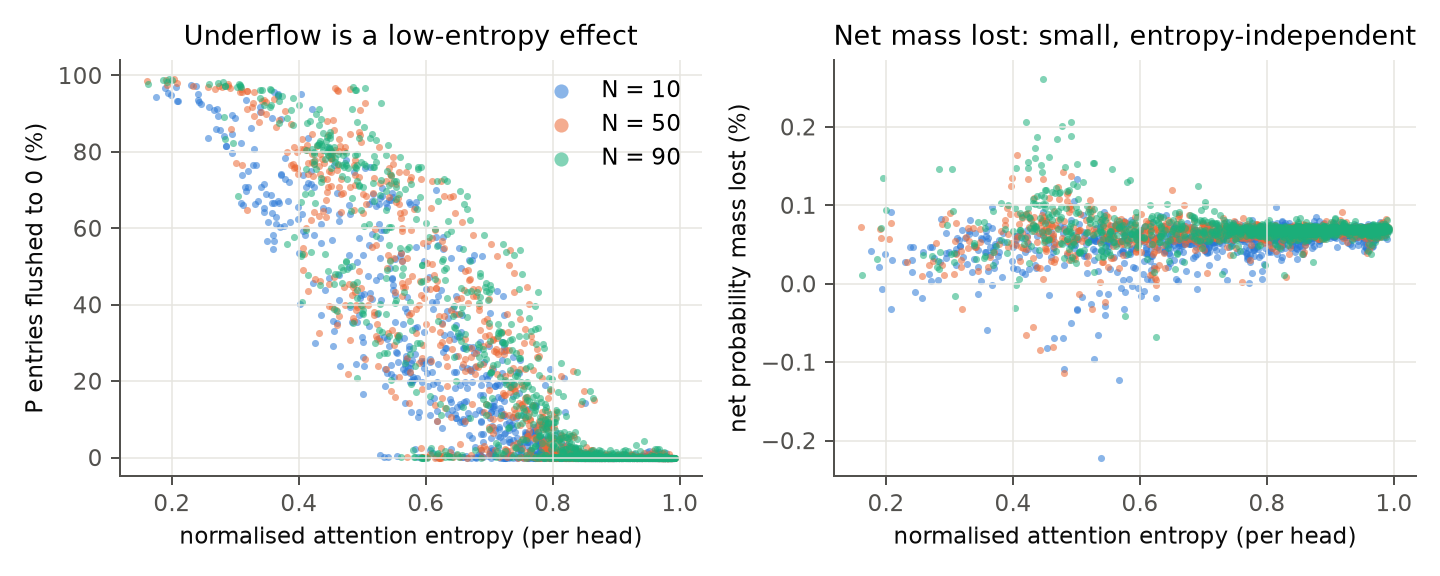}\hfill
\includegraphics[height=4.2cm]{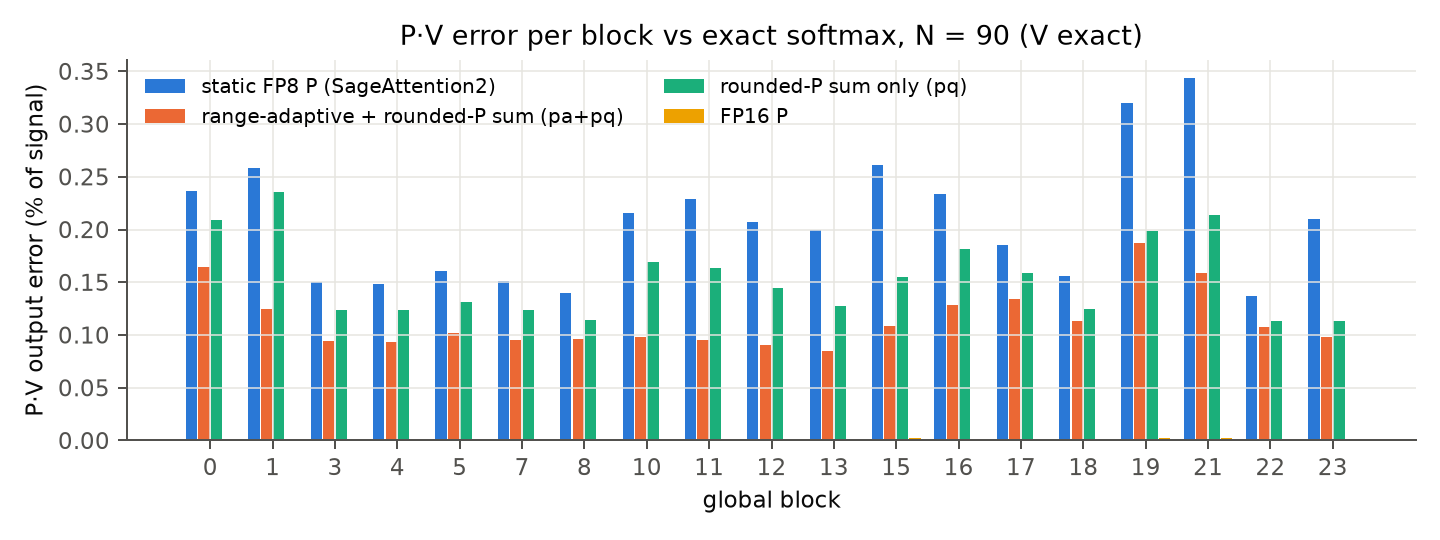}\hfill
\includegraphics[height=4.2cm]{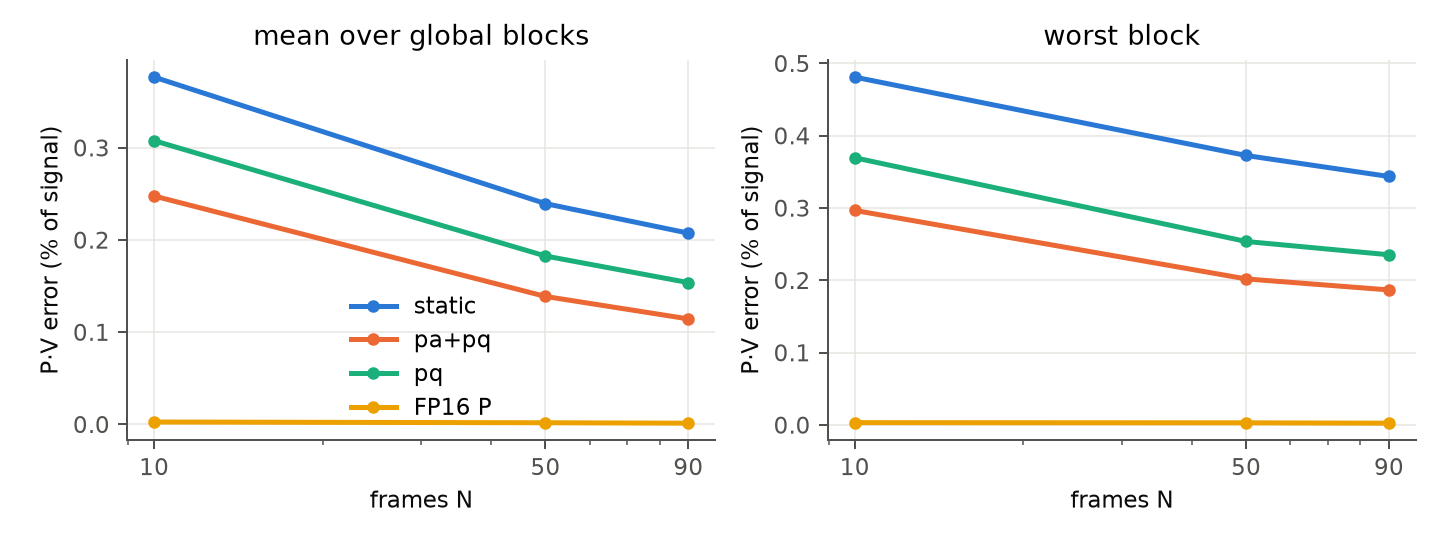}
\caption{FP8 $P$ in \vggto. Left: flushed fraction and lost mass per head against normalised entropy. Middle: per-block $P{\cdot}V$ error at $N{=}90$ for the static cast, \texttt{pq}, \texttt{pa+pq} and FP16 $P$. Right: mean and worst-block error against $N$.}
\label{fig:s-fp8}
\end{figure*}

We emulate the $P{\cdot}V$ step of our kernel (64-key blocks, online running max $m_j$, $P{=}448\,e^{s-m_j}$ cast to E4M3) on 512 query rows $\times$ 16 heads of every global block, with exact FP32 logits and $V$, on five scenes at $N{=}10$, 50 and 90 (Official \vggto runs out of memory at 200). \Cref{fig:s-fp8} shows that flushing to zero happens in \emph{selective} rows (Spearman $\rho{=}{-}0.88$ with entropy), carries negligible mass, and that the relative error falls with $N$. Scaling each 64-key block by its own maximum (``pa'') and accumulating the row sum from the rounded $P$ (``pq'') together halve the error (0.21\%$\to$0.11\% at $N{=}90$); FP16 $P$ is below 0.01\% everywhere. End to end over six datasets $\times$ three seeds, uniform \texttt{pq} has mean $\Delta$AUC@3 $+0.05\pm0.05$ and worst $-0.62\pm0.20$ (\ra: $-0.14\pm0.08$ / $-0.51\pm0.11$) at 1811\,ms vs 1736\,ms ($N{=}50$) and 13.7\,s vs 12.5\,s ($N{=}200$).

\section{Searched vs fixed schedules}
\begin{figure*}[t]
\centering
\includegraphics[width=\linewidth]{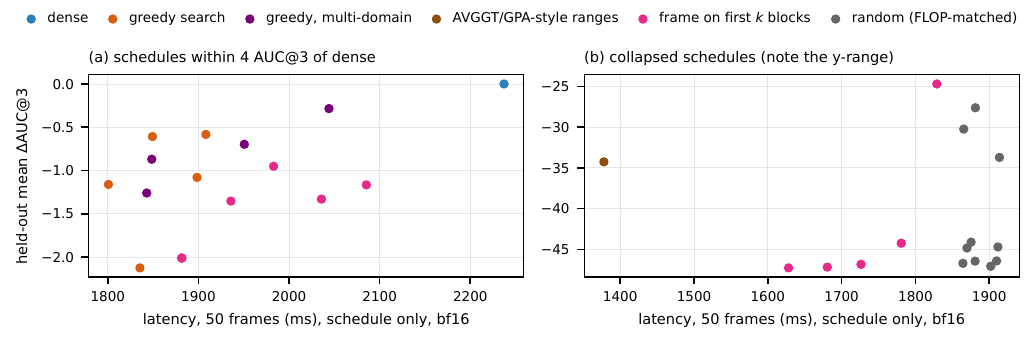}
\caption{Searched vs fixed schedules (schedule only, bf16, PyTorch accuracy). (a) Schedules within 4 AUC@3 of dense. (b) Schedules that collapse; all of them touch a matching or anchoring block.}
\label{fig:s-fixed}
\end{figure*}
\Cref{fig:s-fixed} compares searched schedules with fixed ones. Frame attention on the first $k$ global layers is competitive up to $k{=}7$ (blocks 0--8, $-2.0$ mean) and collapses for $k{\ge}8$ ($-25$ to $-47$). Ten random schedules matched to the searched schedule's attention FLOPs all lose 28--47 AUC@3; every one of them touches a block in 10--23. The AVGGT/GPA-style layer split (frame on 0--8, $2{\times}2$ merge from 10) loses 34.3 on average.

\section{Equal-time accuracy}
\begin{table}[t]
\centering\small\setlength{\tabcolsep}{3pt}
\caption{Extra context frames do not improve pose accuracy on 7-Scenes (7-Scenes-long, 2 sequences; AUC@3 on the same 50 test frames, context frames added evenly over the sequence).}
\label{tab:s-equal}
\resizebox{\linewidth}{!}{\begin{tabular}{@{}llrrr@{}}
\toprule
GPU & method & $N$ & s & AUC@3 \\
\midrule
4070 SUPER & Official & 50 / 70 / 90 & 4.3 / 7.0 / 10.2 & 34.7 / 34.2 / 34.9 \\
 & the exact engine, \ra & 50 & 2.7, 2.0 & 34.6, 34.8 \\
\midrule
Blackwell & Official & 50 / 150 / 340 / 810 & 1.0 / 4.9 / 19.6 / 97.5 & 35.1 / 34.0 / 33.5 / 32.3 \\
 & \ra & 50 / 210 / 460 / 1000 & 0.5 / 4.8 / 19.7 / 85.2 & 34.6 / 33.8 / 32.8 / 31.9 \\
\bottomrule
\end{tabular}}
\end{table}
We tested whether a faster model is more accurate at equal time because it can see more frames (\cref{tab:s-equal}). On 7-Scenes it is not: for every method, AUC@3 on the fixed test frames stays flat or falls as context frames are added. The speed-up is therefore useful for throughput and for sequences Official \vggto cannot fit, not for pose accuracy on this benchmark. One dataset of indoor handheld video; the test frames already cover the whole sequence.

\section{Full evaluation results}
\begin{table*}[t]
\centering\small
\setlength{\tabcolsep}{3pt}
\caption{Evaluation AUC@3 / AUC@30, seed 0, six datasets; native engine. Worst $\Delta$ against the exact engine.}
\label{tab:s-seed0}
\resizebox{\linewidth}{!}{\begin{tabular}{@{}lcccccc|cc@{}}
\toprule
 & ETH3D & CO3D & RE10K & 7-Sc. & NRGBD & TUM-dyn & worst $\Delta_3$ & worst $\Delta_{30}$ \\
\midrule
Exact engine & 49.91/91.05 & 52.85/91.33 & 38.09/77.79 & 22.63/86.08 & 85.93/98.08 & 40.65/84.88 & 0 & 0 \\
uniform SageAttention2 & 50.09/91.28 & 52.06/91.14 & 37.04/76.95 & 21.44/85.76 & 83.79/97.86 & 40.19/84.03 & $-2.14$ & $-0.85$ \\
uniform SageAttention2++ & 49.46/90.68 & 52.30/91.22 & 37.76/77.46 & 22.39/85.99 & 85.60/98.02 & 40.28/83.76 & $-0.56$ & $-1.12$ \\
uniform 16-bit $PV$ & 50.43/91.11 & 52.74/91.33 & 37.88/77.81 & 22.63/86.10 & 85.76/98.09 & 40.56/84.82 & $-0.21$ & $-0.06$ \\
\ra & 49.69/91.07 & 52.46/91.29 & 37.99/77.67 & 22.76/86.16 & 85.51/98.02 & 40.74/85.04 & $-0.41$ & $-0.12$ \\
\quad 16-bit on 15 only & 50.20/91.03 & 52.98/91.37 & 37.74/77.41 & 23.05/86.15 & 85.76/98.09 & 41.20/85.15 & $-0.35$ & $-0.38$ \\
\quad 16-bit on 4,5,8,15 & 50.37/91.04 & 52.70/91.33 & 37.95/77.54 & 23.17/86.10 & 85.68/98.07 & 41.11/85.46 & $-0.25$ & $-0.26$ \\
uniform SageAttention2++ + FP8 GEMMs & 48.95/90.81 & 52.17/91.15 & 37.08/77.10 & 22.39/85.94 & 85.60/98.03 & 39.72/83.65 & $-1.01$ & $-1.23$ \\
\midrule
\raf & 49.57/89.85 & 52.63/91.54 & 38.13/78.18 & 21.89/85.88 & 85.19/97.90 & 40.56/84.76 & $-0.74$ & $-1.20$ \\
greedy search, budget $-2$ & 48.89/90.75 & 52.87/91.51 & 37.47/77.92 & 21.65/85.86 & 84.28/97.66 & 39.26/84.11 & $-1.65$ & $-0.77$ \\
earlier schedule point & 49.69/90.01 & 53.24/91.57 & 37.82/78.11 & 21.52/85.85 & 85.10/97.89 & 40.37/84.66 & $-1.11$ & $-1.04$ \\
multi-dataset schedule + uniform 16-bit $PV$ & 49.91/90.41 & 52.67/91.47 & 37.76/78.03 & 21.81/85.87 & 85.43/97.98 & 40.74/84.56 & $-0.82$ & $-0.64$ \\
multi-dataset schedule + SageAttention2++ & 50.14/90.60 & 52.41/91.39 & 36.67/77.16 & 21.03/85.72 & 82.72/97.72 & 38.89/83.19 & $-3.21$ & $-1.69$ \\
multi-dataset schedule + SageAttention2 + FP8 & 48.83/89.48 & 52.76/91.32 & 36.86/77.51 & 21.23/85.77 & 83.70/97.82 & 38.80/83.03 & $-2.22$ & $-1.85$ \\
outdoor schedule (v1) + SageAttention2 + FP8 & \multicolumn{6}{c|}{$\Delta_3$: $-2.22$ / $-0.81$ / $-0.27$ / $-1.44$ / $-3.13$ / $-0.93$} & $-3.13$ & -- \\
outdoor schedule (v1), bf16 & \multicolumn{6}{c|}{$\Delta_3$: $-0.97$ / $-0.20$ / $-0.19$ / $-1.03$ / $-1.73$ / $-0.83$} & $-1.73$ & -- \\
\bottomrule
\end{tabular}}
\end{table*}
\Cref{tab:s-seed0} gives every seed-0 configuration on the six evaluation datasets. Threshold schedule (merge variant, no search) seed-0 cells (AUC@3/AUC@30): ETH3D 50.09/89.82, CO3D 52.72/91.41, RE10K 38.97/78.08, 7-Scenes 22.26/86.07, NRGBD 85.84/98.00, TUM-dynamic 40.74/85.01.

\paragraph{Official vs the exact engine (parity).}
On the E1.1 splits (3 seeds, ETH3D official frames), Official / the exact engine AUC@3 is: ETH3D 50.7 / 49.9, 7-Scenes 21.3$\pm$1.7 / 21.5$\pm$1.6, NRGBD 85.8$\pm$2.9 / 85.5$\pm$2.3, TUM 39.4$\pm$1.2 / 39.9$\pm$0.9, CO3D 48.2$\pm$0.3 / 48.3$\pm$0.6, RE10K 43.6$\pm$0.8 / 43.6$\pm$0.6. AUC@30 differs by at most 0.3. Depth $\delta_{1.25}$ is identical on ETH3D (99.8), 7-Scenes (94.6) and NRGBD (99.7), and AbsRel differs by at most 0.001, for every configuration in that study including all schedules. (The CO3D/RE10K splits of E1.1 differ from the calibration-disjoint splits of Tab.~1 of the paper, which is why the main table has no official CO3D/RE10K entry.)

\section{Accuracy at 50, 100 and 200 frames}
\Cref{tab:s-longN} gives the five seeded evaluation datasets at $N{=}50$, 100 and 200 (three seeds, lock held). At $N{=}50$, Official \vggto scores within $0.33\pm0.08$ worst-dataset AUC@3 of the exact engine; it does not run 100 frames on 12\,GB. ETH3D (official frames) at $N{=}50$ is in \cref{tab:s-eth50}. On 7-Scenes sequences of 100 frames at $512{\times}384$ (three seeds), the exact engine / \ra / \raf / threshold schedule score 33.29 / 33.19 / 32.34 / 33.98 AUC@3.
\begin{table*}[t]
\centering\small\setlength{\tabcolsep}{3pt}
\caption{AUC@3 / AUC@30 (mean over three seeds) with $N$ frames per sequence, and worst-dataset $\Delta$AUC@3 against the exact engine (mean\,$\pm$\,std, ddof${=}1$, over seeds). $^\dagger$Post hoc (selected after the main evaluation; \cref{sec:s-ctrl}); its RE10K loss at 100/200 frames is the largest of any schedule we tested.}
\label{tab:s-longN}
\resizebox{\linewidth}{!}{\begin{tabular}{@{}llccccc|c@{}}
\toprule
$N$ & & 7-Scenes & NRGBD & TUM-dyn & CO3D & RE10K & worst $\Delta_3$ \\
\midrule
50 & Official \vggto & 21.5 / 86.2 & 87.1 / 98.1 & 40.1 / 84.0 & 59.4 / 94.0 & 44.8 / 84.9 & $+0.00${\scriptsize$\pm$0.09} \\
 & Exact engine & 21.4 / 86.2 & 86.8 / 98.1 & 39.9 / 84.0 & 59.3 / 94.0 & 44.7 / 84.8 & 0 \\
 & \ra & 21.4 / 86.2 & 86.6 / 98.1 & 39.9 / 84.0 & 59.3 / 93.9 & 44.6 / 84.8 & $-0.32${\scriptsize$\pm$0.14} \\
 & \raf & 21.0 / 86.0 & 86.1 / 98.1 & 39.8 / 83.6 & 58.9 / 93.9 & 44.4 / 86.0 & $-0.65${\scriptsize$\pm$0.14} \\
 & threshold schedule & 21.3 / 86.1 & 86.6 / 98.2 & 39.9 / 84.1 & 58.8 / 93.9 & 44.7 / 86.2 & $-0.62${\scriptsize$\pm$0.21} \\
 & sensitivity-only greedy$^\dagger$ & 21.2 / 86.1 & 86.9 / 98.2 & 39.9 / 83.9 & 58.9 / 93.9 & 44.0 / 86.0 & $-0.81${\scriptsize$\pm$0.43} \\
\midrule
100 & Exact engine & 26.3 / 87.1 & 87.5 / 98.3 & 41.8 / 83.7 & 41.6 / 90.6 & 56.7 / 93.8 & 0 \\
 & \ra & 26.3 / 87.1 & 87.5 / 98.3 & 41.8 / 83.7 & 41.5 / 90.6 & 56.7 / 93.8 & $-0.13${\scriptsize$\pm$0.01} \\
 & \raf & 25.9 / 86.9 & 86.5 / 98.1 & 42.1 / 83.4 & 41.8 / 90.6 & 55.5 / 93.5 & $-1.20${\scriptsize$\pm$0.26} \\
 & threshold schedule & 26.6 / 87.1 & 87.2 / 98.2 & 41.9 / 83.8 & 41.0 / 90.5 & 54.8 / 93.5 & $-1.87${\scriptsize$\pm$0.35} \\
 & sensitivity-only greedy$^\dagger$ & 26.1 / 87.0 & 87.3 / 98.3 & 41.9 / 83.5 & 41.7 / 90.5 & 53.3 / 93.1 & $-3.42${\scriptsize$\pm$0.46} \\
\midrule
200 & Exact engine & 26.3 / 87.2 & 86.3 / 98.1 & 42.0 / 84.5 & 58.8 / 94.6 & 50.7 / 93.5 & 0 \\
 & \ra & 26.2 / 87.2 & 86.3 / 98.1 & 41.8 / 84.4 & 58.7 / 94.6 & 50.5 / 93.5 & $-0.22${\scriptsize$\pm$0.01} \\
 & \raf & 25.9 / 87.0 & 85.3 / 98.0 & 42.1 / 84.1 & 58.7 / 94.6 & 50.0 / 93.4 & $-1.03${\scriptsize$\pm$0.16} \\
 & threshold schedule & 26.3 / 87.1 & 85.9 / 98.1 & 41.9 / 84.6 & 58.6 / 94.6 & 49.2 / 93.3 & $-1.50${\scriptsize$\pm$0.24} \\
 & sensitivity-only greedy$^\dagger$ & 26.1 / 87.1 & 86.2 / 98.1 & 41.9 / 84.4 & 59.6 / 94.7 & 48.6 / 93.2 & $-2.10${\scriptsize$\pm$0.14} \\
\bottomrule
\end{tabular}}
\end{table*}

\section{Calibration and evaluation data}
\Cref{tab:s-split} lists calibration and evaluation data per dataset. Evaluation is sequence-disjoint from calibration everywhere and dataset-disjoint for ETH3D and NRGBD only. Per-seed frame maps of every evaluation sequence will be released with the code.
\begin{table*}[t]
\centering\small
\caption{Calibration vs evaluation data. ``Disjoint'': level at which the evaluation data is disjoint from calibration.}
\label{tab:s-split}
\begin{tabular}{@{}llll@{}}
\toprule
dataset & calibration (sequences $\times$ frames) & evaluation (sequences $\times$ frames, seeds 0--2) & disjoint \\
\midrule
Mip-NeRF~360 & bicycle, garden, stump, kitchen ($4\times50$) & -- (NVS only, \cref{tab:s-nvs}) & -- \\
CO3Dv2 & banana, donut ($10\times10$) & 40 sequences of 8 other categories ($\times10$) & sequence + category \\
RealEstate10K & 30 sequences ($\times10$) & 38 other sequences ($\times10$) & sequence \\
7-Scenes & first training sequence per scene ($7\times50$) & 10 test sequences ($\times10$) & sequence \\
TUM-dynamic & fr3 walking xyz/halfsphere/rpy, sitting xyz \emph{validation} ($4\times50$) & 8 other takes ($\times10$) & sequence \\
ETH3D & -- & 13 scenes, official frames & dataset \\
NRGBD & -- & 9 scenes ($\times10$) & dataset \\
\bottomrule
\end{tabular}
\end{table*}
Results restricted to the two dataset-disjoint datasets (ETH3D, NRGBD): \ra $-0.23$/$-0.08$, \raf $-0.34$/$-0.49$, threshold schedule $+0.17$/$-0.36$ mean $\Delta$AUC@3 over three seeds (Tab.~1 of the paper).

\paragraph{Observation on seed spread.} On the evaluation data, the seed standard deviation of the worst-dataset $\Delta$AUC@3 is $\pm0.54$ (ddof${=}1$, as everywhere in this supplement) for threshold schedule and $\pm0.17$ for \raf. This was observed after the operating points were fixed and did not influence any choice.

\section{Search details}
\paragraph{Precision search.} Single-layer gains (worst-dataset $\Delta_3$ improvement $+0.5\times$ mean improvement) when one layer of uniform SageAttention2++ gets 16-bit $PV$: block 15 $+0.88$, 4 $+0.69$, 5 $+0.56$, 8 $+0.39$, 17 $+0.16$, 11 $+0.15$, 16 $+0.11$, 21 $+0.10$, 10 $+0.09$, 19 $+0.07$, 12 $+0.06$, 22/23 $+0.05$, 18 $-0.01$, 7 $-0.05$, 3 $-0.18$, 13 $-0.20$, 1 $-0.56$, 0 $-1.00$. Cumulative calibration worst $\Delta_3/\Delta_{30}$: base $-1.37/-0.83$; +15: $-0.67/-0.07$; +4: $-0.30/-0.05$ (budget $-0.5/-0.3$ met). The search ran with 16-bit $PV$ at FP32 accumulation; the deployed kernel accumulates in FP16 per 64-key block (same kernel error, rel.\ L2 0.0028 vs FP64, 24\% faster), and every evaluation number re-measures the deployed kernel.
\paragraph{Calibration datasets} (uniform precisions, calibration $\Delta$AUC@3 outdoor/object/indoor/RGB-D/dynamic): SageAttention2 $-0.27/-0.15/-0.54/-0.23/-0.61$; SageAttention2++ $+0.35/-0.37/-0.59/-0.31/-1.37$; 16-bit $PV$ (FP32 acc.) $+0.07/+0.07/-0.07/-0.05/+0.12$; bf16 + FP8 GEMMs $-0.81/-0.81/-0.84/-0.44/-0.86$.
\paragraph{Schedule search at fixed precision.} Single-layer calibration worst $\Delta_3$ with the precision of \ra: block 0 merge$_4$ $-0.02$ / frame $-0.75$; 3 merge$_4$ $-0.32$; 4 frame $-0.47$; 5 frame $-0.32$; 8 and 19 merge$_4$ $-0.32$; 22 merge$_4$ $-0.91$; frame on any of blocks 10--23 $-18.8$ to $-78.0$. Budgets $-0.5$, $-1$, $-2$ return 2, 5 and 8 changed layers (\cref{tab:s-names}); calibration worst $-0.48$, $-0.84$, $-1.99$. Cost of the search for \raf: 38 single-layer evaluations plus 11 greedy trials (49 configurations, each on five calibration datasets); with the budget sweep, 38 plus 23 unique greedy configurations.
\paragraph{Re-check on fresh calibration frames.} \Cref{tab:s-recheck} repeats the cumulative check (five calibration datasets, against the exact engine) on the search's frames (seed 0) and two new samples.
\begin{table}[t]
\centering\small\setlength{\tabcolsep}{2.4pt}
\caption{Cumulative calibration check against the exact engine, worst over five calibration datasets, $\Delta$AUC@3 / $\Delta$AUC@30 per sample (weakest dataset). Operator budget: $-1$ / $-0.5$. All cells use seed-matched calibration frames; $^\ast$the first 3-sample run reused seed-0 outdoor frames for seeds 1 and 2 during its search (\cref{sec:s-sens}).}
\label{tab:s-recheck}
\resizebox{\linewidth}{!}{\begin{tabular}{@{}lccc@{}}
\toprule
 & seed 0 (search) & seed 1 & seed 2 \\
\midrule
\ra & $-0.44$ / $-0.12$ & $-0.18$ / $-0.39$ & $-0.67$ / $-0.02$ \\
\raf & $-0.84$ / $-0.35$ & $\mathbf{-1.26}$ / $\mathbf{-0.68}$ (object, indoor) & $-1.00$ / $-0.28$ \\
threshold schedule & $\mathbf{-1.42}$ / $-0.33$ (outdoor) & $\mathbf{-1.52}$ / $\mathbf{-0.99}$ & $\mathbf{-1.73}$ / $-0.20$ (outdoor) \\
3-sample, first run$^\ast$ & $-0.78$ / $-0.34$ & $-0.80$ / $-0.39$ & $-0.81$ / $-0.20$ \\
3-sample, seed-matched & $-0.79$ / $-0.18$ & $-0.99$ / $-0.39$ & $-0.84$ / $-0.11$ \\
\bottomrule
\end{tabular}}
\end{table}
\begin{table}[t]
\centering\small\setlength{\tabcolsep}{2.5pt}
\caption{Calibration set size and dataset (greedy, outdoor protocol, budget $-1/-0.5$). Evaluation worst $\Delta$AUC@3 (PyTorch; Nine@50, ETH3D, 7-Scenes). Jaccard: overlap of (layer, operator) pairs with the 4-scene schedule.}
\label{tab:s-calib}
\begin{tabular}{@{}lrrr@{}}
\toprule
calibration & evals & worst eval. & Jaccard \\
\midrule
1 scene $\times$ 10 & 38 & $-35.35$ & 0.24 \\
2 scenes $\times$ 10 & 38 & $-11.77$ & 0.27 \\
4 scenes $\times$ 10 & 47 & $-1.12$ & 0.36 \\
8 seqs $\times$ 10 (3 datasets) & 44 & $-2.62$ & 0.40 \\
1 scene $\times$ 50 & 37 & $-29.37$ & 0.27 \\
2 scenes $\times$ 50 & 41 & $-9.25$ & 0.31 \\
4 scenes $\times$ 50 & 47 & $-1.37$ & 1.00 \\
object only (10$\times$10) & 39 & $-10.27$ & 0.29 \\
indoor only (10$\times$10) & 39 & $-17.16$ & 0.27 \\
\bottomrule
\end{tabular}
\end{table}
\paragraph{Visit order.} Greedy on outdoor calibration with three orders: by single-layer frame deviation (worst eval. $-1.37$, 1533\,ms with SageAttention2+FP8), by merge$_2$ deviation ($-0.91$, 1544\,ms), by latency gain per AUC lost ($-1.03$, 1514\,ms). Of 35 local swaps around the first schedule that add block 12 merge$_2$, only one stays within calibration budget, and it costs $-4.28$ on Nine@50.
\paragraph{Non-additivity.} \Cref{tab:s-nonadd} lists single-layer and in-context (marginal) costs.
\begin{table}[t]
\centering\small\setlength{\tabcolsep}{2.5pt}
\caption{Single vs marginal $\Delta$AUC@3 (PyTorch) for the layers of the v1 schedule. Marginal: AUC(schedule) $-$ AUC(schedule without the layer).}
\label{tab:s-nonadd}
\resizebox{\linewidth}{!}{\begin{tabular}{@{}lrrrr@{}}
\toprule
 & \multicolumn{2}{c}{single} & \multicolumn{2}{c}{marginal}\\
layer:op & Cal-out & eval. & Cal-out & eval. \\
\midrule
0:merge4 & $+0.04$ & $-0.17$ & $+0.14$ & $+0.19$ \\
1:merge4 & $+0.08$ & $-0.22$ & $+0.04$ & $-0.09$ \\
3:frame & $-1.50$ & $-0.65$ & $-0.25$ & $-0.50$ \\
4:frame & $-0.25$ & $-0.16$ & $+0.01$ & $-0.13$ \\
5:frame & $+0.05$ & $-0.50$ & $+0.70$ & $-0.00$ \\
7:merge4 & $-0.27$ & $-0.41$ & $-0.20$ & $-0.46$ \\
19:merge4 & $-0.11$ & $-0.20$ & $+0.10$ & $-0.11$ \\
12:merge2 & $+0.01$ & $-2.16$ & $-1.05$ & $-1.73$ \\
\midrule
sum of singles / measured, ETH3D & \multicolumn{4}{c}{$-3.13$ / $-1.37$} \\
sum of singles / measured, 7-Scenes & \multicolumn{4}{c}{$-2.80$ / $-1.03$} \\
\bottomrule
\end{tabular}}
\end{table}

\section{Operator ablations}
\paragraph{Merge factor, bias, merged queries} (merged layers of the v1 schedule; worst eval. $\Delta$AUC@3): merge$_2$ $+\log c$ $-1.23$ / no bias $-1.32$; merge$_4$ $+\log c$ $-1.37$ / no bias $-1.78$; merge$_8$ $+\log c$ $-1.25$ / no bias $-1.94$; merging Q as well: $-3.05$ (merge$_2$), $-4.24$ (merge$_4$). The bias is worth 0.5--0.9 on Nine@50.
\paragraph{Special tokens and frame 0.} Merging the other frames' special tokens into one mean token: $-1.71$; dropping them: $-2.34$ (vs $-1.37$). Cross-frame exchange through special tokens alone instead of merged K/V: $-25.2$. Turning off the frame-0 exemption costs 4.7 on ETH3D for the v1 schedule but nothing for the multi-dataset schedule ($-0.78$ vs $-1.03$).
\paragraph{Register-only blocks.} Running blocks 2, 6, 9, 14, 20 over all tokens (frame or global) drives AUC to 0 on every set: they are not a speed lever.
\paragraph{Transfer across $N$} (native, nine COLMAP scenes, v1 / multi-dataset schedules with SageAttention2+FP8, $\Delta$AUC@3 vs the exact engine): at 10/50/100/200 frames $-1.81/-1.50/-1.87/-2.85$ and $-0.50/-0.71/-1.14/-1.71$. A 50-frame calibration transfers to 10--100 frames; expect about one extra point at 200 (5 scenes at $N{=}200$).

\section{Kernels}
\begin{table}[t]
\centering\small\setlength{\tabcolsep}{2.5pt}
\caption{Global attention, 16 heads $\times$ 64, $N\times689$ tokens, random bf16 inputs; ms (dense-equivalent TFLOPS). RTX 4070 SUPER, lock held.}
\label{tab:s-kern}
\begin{tabular}{@{}lrr@{}}
\toprule
kernel & $N{=}50$ & $N{=}200$ \\
\midrule
ours, INT8/FP8, FP16 acc. & 22.9 (203) & 376.6 (204) \\
ours, INT8/FP8, FP32 acc. & 29.5 (159) & 445.4 (173) \\
SageAttention2 (official, fp32+fp32) & 30.9 (157) & 474.5 (164) \\
SageAttention2++ (official, fp32+fp16) & 27.2 (179) & 414.0 (188) \\
SageAttention 1.0.6 & 37.2 (131) & 580.0 (134) \\
SDPA FlashAttention-2 (bf16) & 69.6 (70) & 1101.3 (71) \\
SDPA cuDNN (bf16) & 66.8 (73) & 1059.5 (73) \\
ours merge$_4$, FP16 acc. & 2.7 & 41.9 \\
\bottomrule
\end{tabular}
\end{table}
\Cref{tab:s-kern} compares attention kernels. At \vggto shapes our FP8 GEMM (per-token activation, per-channel weight scales) reaches 140--152 TFLOPS against 137--144 for PyTorch \texttt{\_scaled\_mm} with tensor-wise scales and 39--54 with row-wise scales; cuBLASLt bf16 reaches 67--76. Our bf16 frame attention reaches 67/83 TFLOPS (FP32/FP16 accumulation) against 51 for SDPA FlashAttention-2. Per layer at $N{=}50$ including the quantisation prologue: FP8/FP32 acc.\ 29.0\,ms, FP8/FP16 acc.\ 23.8, 16-bit/FP32 acc.\ 47.5, 16-bit/FP16 acc.\ 36.1. Relative L2 error vs FP64 on synthetic inputs: 0.0040 (FP8 $PV$), 0.0028 (16-bit $PV$).

\paragraph{Quantised attention kernel.} Head dimension 64. $QK^\top$: keys are smoothed by subtracting a mean key as in SageAttention (the per-channel mean over the whole sequence, per batch and head; a constant shift of each logit row, hence exact); $Q$ is quantised to INT8 with one scale per 16 query rows and $K$ with one scale per 64 keys, and products use INT8 \texttt{mma.m16n8k32} with INT32 accumulation. The $+\log c_j$ bias of merged keys is added to the scores in registers before the online-softmax row maximum is taken, so it enters the INT8 path only after the integer product. $P{\cdot}V$: probabilities are formed relative to the running row maximum and cast to FP8 E4M3; $V$ is scaled to $\pm2.25$ and cast to E4M3; FP8 \texttt{mma} accumulates in FP16 over each 64-key block, folded into FP32. The 16-bit path keeps $P$ and $V$ in FP16 ($V$ scaled to $\pm256$) with the same two-level accumulation. Each CTA has 4 warps and 128 query rows (two 16-row m-tiles per warp); keys stream through a 3-stage ring of 64-key tiles. Merged keys and values are pooled after QK-norm; frame attention reuses the kernels with a $[BN,T]$ layout.

\section{Memory}
\begin{figure}[t]
\centering
\includegraphics[width=\linewidth]{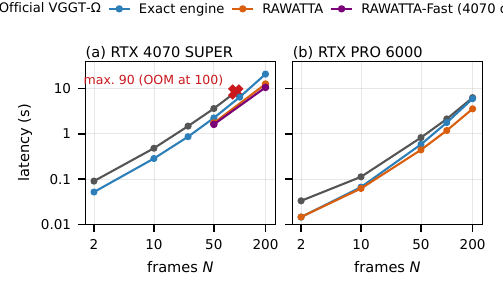}
\vspace{-1.8em}
\caption{\textbf{Latency vs frames} (kitchen, $512{\times}336$, full forward, exclusive GPU). (a) Official \vggto runs out of memory at 100 frames on 12\,GB. (b) On Blackwell, the exact engine alone is barely faster than Official at 200 frames; the gain comes from the method.}
\label{fig:scaling}
\end{figure}
\begin{table*}[t]
\centering\small\setlength{\tabcolsep}{3pt}
\caption{7-Scenes-long (2 sequences, $512{\times}384$), RTX 4070 SUPER 12\,GB: peak GiB / latency s / AUC@3. Ports keep Official's fp32 weights. The schedule row is the multi-dataset schedule with SageAttention2+FP8 (after the duplicate-weight fix). $^\ast$\ra/earlier-schedule cells ran while the GPU was shared: latency not reported, $N{=}200$ failed for lack of free memory (at $512{\times}336$ \ra runs 200 frames at a 9.7\,GiB engine peak, locked session).}
\label{tab:s-mem}
\begin{tabular}{@{}lccccc@{}}
\toprule
method & $N{=}10$ & 50 & 100 & 200 & 500 \\
\midrule
Official \vggto & 6.49 / 0.571 / 31.1 & 8.75 / 4.32 / 35.1 & OOM & OOM & OOM \\
SparseVGGT & 6.49 / 0.558 / 30.4 & 8.75 / 3.63 / 34.9 & OOM & OOM & OOM \\
GPA-VGGT (1B config) & 6.49 / 0.551 / 30.0 & 8.75 / 3.23 / 34.4 & OOM & OOM & OOM \\
FastVGGT & 6.49 / 0.565 / 28.5 & 8.75 / 3.29 / 33.5 & OOM & OOM & OOM \\
chunked \vggto (60/30) & = Official & = Official & 9.55 / 17.0 / 29.0 & 9.70 / 34.0 / 28.4 & 10.14 / 90.6 / 24.2 \\
Exact engine & 3.23 / 0.339 / 31.5 & 4.82 / 2.71 / 35.2 & 6.91 / 7.98 / 33.2 & 10.63 / 26.1 / 32.8 & OOM (plan 20.8\,GB) \\
multi-dataset schedule + Sage2 + FP8 & -- & 4.66 / 1.85 / 33.6 & 6.76 / 4.69 / 31.7 & 10.48 / 13.3 / 31.6 & OOM \\
\ra & -- & 4.98 / -- / 35.0 & 7.23 / -- / 33.2 & not run$^\ast$ & -- \\
earlier schedule point & -- & 5.04 / -- / 34.2 & 7.35 / -- / 31.9 & not run$^\ast$ & -- \\
\bottomrule
\end{tabular}
\end{table*}
\Cref{tab:s-mem} gives the memory study. Planner limits for a single pass on 12\,GB: about 229 frames at $512{\times}384$ and 259 at $512{\times}336$ (upper bounds: the exact engine at $512{\times}384$, $N{=}220$, passed the planner but failed to allocate). \emph{Planner ablation} (kitchen, $512{\times}336$, fixed schedule, $N{=}200$): forced offload brings the exact engine from 9.98 to 4.89\,GB at $+5.5\%$ latency; for the balanced schedule the planner's offload gives 5.76\,GB against 11.42\,GB without it, at $+11.5\%$ latency. The planner, not the schedule, accounts for the whole saving.

\section{Second GPU class (RTX PRO 6000 Blackwell)}
\begin{table*}[t]
\centering\small\setlength{\tabcolsep}{3pt}
\caption{RTX PRO 6000 Blackwell Server Edition (sm\_120, 96\,GB, CUDA 13.0, torch 2.11). Kitchen $512{\times}336$ latency ms (peak MiB), and 7-Scenes-long AUC@3 / latency ms.}
\label{tab:s-gpu2}
\begin{tabular}{@{}lccccc|ccc@{}}
\toprule
 & $N{=}2$ & 10 & 50 & 100 & 200 & 7S $N{=}50$ & 7S $N{=}200$ & 7S $N{=}500$ \\
\midrule
Official & 33.4 (4601) & 112.9 & 826.6 (6923) & 2130.3 & 6274.9 (14550) & 35.14 / 997 & 32.92 / 7851 & 31.97 / 39322 \\
Exact engine & 14.6 (2470) & 67.0 & 585.2 (4614) & 1773.4 & 5953.6 (9830) & 34.73 / 731 & 32.73 / 7608 & 31.89 / 42573 \\
uniform SageAttention2 & 14.6 & 63.0 & 435.2 & 1152.0 & 3442.7 (10278) & 34.16 / 525 & -- & -- \\
\ra & 14.7 & 62.5 & 443.4 (4742) & 1184.6 & 3563.4 (10406) & 34.56 / 534 & 32.60 / 4470 & 31.88 / 22928 \\
earlier schedule point & 14.6 & 60.8 & 398.5 (4806) & 1002.2 & 2846.0 (10630) & 33.88 / 476 & -- & -- \\
\bottomrule
\end{tabular}
\end{table*}
\Cref{tab:s-gpu2} has the full Blackwell numbers (card exclusive, one fresh process per cell, no lock). \ra is 1.86\x and 1.76\x faster than Official at 50 and 200 frames, the exact engine only 1.41\x and 1.05\x: on this card the gain is the method's. Our kernels ran as compute\_89 PTX compiled by the driver (no sm\_120 tuning). Single-pass probes (5-minute cap, no OOM reached): Official 1390 frames (276\,s, 85.0\,GB), the exact engine 1310 (280\,s, 56.3\,GB), \ra 1830 (277\,s, 83.4\,GB). At 500 frames, GPA-VGGT as released (PyTorch) takes 17.4\,s at 32.20 AUC@3 and SparseVGGT 34.7\,s at 31.98, against 22.9\,s and 31.88 for \ra and 39.3\,s and 31.97 for Official (7-Scenes-long).

\section{Generality: VGGT-1B}
We ran the procedure on VGGT-1B in PyTorch (real SageAttention2 kernel via SpargeAttn; FP8 GEMMs simulated, never timed). The sensitivity band is blocks 11--17 (frame: $-33$ to $-78$ AUC@3), not 10--23. Uniform SageAttention2 is near-lossless ($\Delta$AUC@3 $-0.54$ to $+0.80$ on CO3Dv2/RE10K/7-Scenes/ETH3D). In an earlier protocol, distinct from the search of Sec.~5.6 of the paper (PyTorch, budgets on the calibration mean, a different schedule), a safe-budget schedule (46\% of attention FLOPs) is lossless on CO3Dv2 ($-0.1$) and 7-Scenes but costs $-1.3$ on RE10K and $-3.8$ on ETH3D (not a calibration dataset). GPA-VGGT with its own configuration is more accurate on ETH3D (27.41 vs 26.38 AUC@3) and uses less than half the memory; SparseVGGT is most accurate ($-0.3$) and slowest (1.11\x). At 50 frames (PyTorch): Sage-only 1.36\x, safe schedule + Sage 1.49\x, GPA-VGGT 1.39\x over bf16 dense; official dense (fp32 weights) OOMs.
\begin{table*}[t]
\centering\small
\setlength{\tabcolsep}{3pt}
\caption{\textbf{Other models (Sec.~5.6 of the paper).} PyTorch, SageAttention2 kernel of SpargeAttn~\cite{spargeattn}, bf16. First row of each model: AUC@3; other rows: $\Delta$AUC@3 against dense bf16, mean over three seeds (ETH3D: official frames). Worst: per-seed minimum over the six datasets, mean\,$\pm$\,std. Latency $N{=}50$ (Mip-NeRF~360 bicycle, lock held, one session). $N{=}200$ runs out of memory in PyTorch for every configuration of both models.}
\label{tab:s-gen}
\begin{tabular}{@{}llcccccc|cr@{}}
\toprule
 & & CO3D & RE10K & 7-Sc. & NRGBD & TUM-dyn & ETH3D & worst $\Delta_3$ & s \\
\midrule
\multirow{4}{*}{VGGT-1B} & dense bf16 & 59.99 & 15.87 & 11.69 & 78.71 & 28.15 & 30.14 & 0 & 5.85 \\
 & Sage only & $-$0.18 & $-$0.42 & $-$0.03 & $-$0.66 & $+$0.00 & $+$0.34 & $-0.84${\scriptsize$\pm$0.22} & 4.31 \\
 & threshold schedule (\vggto thresholds) & $-$6.24 & $-$3.69 & $-$0.24 & $-$7.60 & $-$7.19 & $-$6.21 & $-7.82${\scriptsize$\pm$0.85} & 3.70 \\
 & searched & $-$0.50 & $-$1.50 & $-$0.19 & $-$3.79 & $-$0.31 & $-$2.50 & $-3.79${\scriptsize$\pm$0.76} & 3.82 \\
\midrule
\multirow{4}{*}{$\pi^3$} & dense bf16 & 61.37 & 43.64 & 11.85 & 79.37 & 20.31 & 36.58 & 0 & 4.88 \\
 & Sage only & $-$0.34 & $+$0.12 & $-$0.16 & $-$0.52 & $+$0.03 & $+$0.23 & $-0.74${\scriptsize$\pm$0.44} & 3.72 \\
 & threshold schedule (\vggto thresholds) & $-$1.71 & $-$1.31 & $-$4.00 & $-$4.45 & $+$1.23 & $-$8.83 & $-8.83${\scriptsize$\pm$0.00} & 3.23 \\
 & searched & $-$0.49 & $-$0.16 & $-$0.27 & $-$0.11 & $-$0.31 & $-$0.74 & $-1.09${\scriptsize$\pm$0.58} & 3.50 \\
\bottomrule
\end{tabular}
\end{table*}
\paragraph{Threshold schedule and search on VGGT-1B and $\pi^3$.} \Cref{tab:s-gen} reports the threshold schedule with \vggto's thresholds on each model's own statistics (merge variant; VGGT-1B: merge$_4$ on 0--9, 16--20, 22, 23; $\pi^3$: merge$_4$ on 0--10 and 17, all frames pooled since $\pi^3$ has no reference frame) and the greedy search with \vggto's protocol: the same five calibration datasets, SageAttention2 everywhere, operators frame and merge$_4$, cumulative budget $-1$ AUC@3 / $-0.5$ AUC@30 against the model's own Sage-only configuration, candidates restricted to the blocks the regime map marks context or anchoring (VGGT-1B: 0--10, 17--23; $\pi^3$: 0--10, 16, 17). It needs 51 (VGGT-1B) and 35 ($\pi^3$) calibration evaluations of about 95\,s. VGGT-1B: frame on 1, 2, 3, 7, merge$_4$ on 5, 6, 8, 18, 20, 22, 23 (57\% of the global-attention FLOPs), calibration worst $-0.91$; $\pi^3$: frame on 3, merge$_4$ on 0, 1, 5, 7 (74\%), calibration worst $-0.78$. Within the candidates, the check rejects the stage boundaries (VGGT-1B 9, 10, 17; $\pi^3$ 6, 8--10, 16, 17) that the map calls context: the map is a candidate filter, the check makes the schedule safe. On VGGT-1B the evaluation cost ($-3.79$, NRGBD) is about 3.5\x the calibration estimate; no calibration dataset covers NRGBD. An earlier VGGT-1B search with outdoor calibration lost 0.1 / 1.3 / 3.8 AUC@3 on CO3D / RE10K / ETH3D (seed 0).

\section{Downstream NVS (per scene)}
We feed \vggto poses and depth to a Gaussian-splatting trainer~\cite{gs3d} (30k iterations, bundle adjustment, voxel-merged depth initialisation, per-image pose refinement, every 8th view held out, seed 0, lock held) on 13 scenes of Mip-NeRF~360, Tanks and Temples and Deep Blending. With \ra poses, the 9 Mip-NeRF~360 scenes reach 27.14\,dB against 27.27 with the exact engine and 27.34 with COLMAP poses and the same trainer, and the pose stage is 1.36\x faster (20.5 vs 27.9\,s); with \raf, 26.99\,dB at 1.54\x (18.1\,s) (\cref{tab:s-nvs}). Photos-to-splat time stays dominated by training. Three cells first collapsed because a few extreme depth outliers of the quantised model inflated the initialisation extent, so that all points fell into a few voxels, while bundle adjustment and the poses were normal; clipping outliers (per-axis 1st--99th percentile box, then a 5\x MAD filter) before voxel merging recovered them (room \raf 19.81$\to$30.65, drjohnson \ra 14.64$\to$19.46, playroom \raf 20.27$\to$23.02\,dB). These three cells use the fix; all other cells ran before it, and on clean scenes the fix changes PSNR by at most $\pm$0.07\,dB (garden, counter). On 12\,GB Official \vggto sees at most 90 of garden's 185 frames (26.29 vs 27.61\,dB), and InstantSplat~\cite{instantsplat} runs out of memory at initialisation. With an earlier FP8-GEMM operating point (``earlier fast point used in NVS'', \cref{tab:s-names}; not one of the paper's points), photos-to-splat was 4.8--7.6\x faster than COLMAP with the same trainer (4 scenes).
\begin{table}[t]
\centering\small\setlength{\tabcolsep}{2pt}
\caption{COLMAP-free NVS, test-view PSNR (pose-stage s). $^\ast$Cell with the outlier-clipped initialisation. T\&T train is poor for every row, including COLMAP; single-seed Deep Blending cells vary by $\pm$3\,dB between configurations.}
\label{tab:s-nvs}
\resizebox{\linewidth}{!}{\begin{tabular}{@{}lcccc@{}}
\toprule
scene & COLMAP + trainer & Exact engine & \ra & \raf \\
\midrule
bicycle & 25.49 & 25.43 (20.6) & 25.34 (16.1) & 25.27 (14.0) \\
bonsai & 32.40 & 32.48 (43.0) & 32.55 (29.6) & 32.39 (25.5) \\
counter & 29.13 & 29.00 (43.0) & 28.94 (29.6) & 28.97 (25.5) \\
flowers & 21.65 & 21.85 (17.0) & 21.99 (13.9) & 22.01 (12.4) \\
garden & 27.77 & 27.61 (19.3) & 27.59 (15.2) & 27.55 (15.7) \\
kitchen & 29.32 & 29.37 (43.0) & 28.74 (29.6) & 29.34 (25.6) \\
room & 31.95 & 32.01 (43.0) & 31.51 (29.6) & 30.65$^\ast$ (25.6) \\
stump & 26.35 & 25.76 (9.9) & 25.86 (9.8) & 25.12 (9.0) \\
treehill & 21.97 & 21.91 (12.1) & 21.69 (11.0) & 21.60 (10.0) \\
truck (T\&T) & 25.80 & 25.51 (33.4) & 25.44 (24.1) & 25.50 (21.1) \\
train (T\&T) & 19.25 & 15.49 (33.0) & 17.87 (24.0) & 17.41 (21.2) \\
drjohnson (DB) & 24.31 & 19.96 (44.5) & 19.46$^\ast$ (29.4) & 20.62 (25.5) \\
playroom (DB) & 27.32 & 23.92 (42.9) & 23.38 (29.4) & 23.02$^\ast$ (25.6) \\
\midrule
mean, Mip-NeRF~360 (9) & 27.34 & 27.27 (27.9) & 27.14 (20.5) & 26.99 (18.1) \\
\quad SSIM / LPIPS & 0.807 / 0.211 & 0.802 / 0.216 & 0.798 / 0.220 & 0.794 / 0.219 \\
mean, all 13 & 26.36 & 25.41 (31.1) & 25.41 (22.4) & 25.35 (19.7) \\
\bottomrule
\end{tabular}}
\end{table}

\section{Additional measurements}
\label{sec:s-v32}
All runs in this section: RTX 4070 SUPER, GPU lock held, frame-sampling seeds 0, 1, 2, $\pm$ = sample std (ddof${=}1$).

\paragraph{ETH3D at 50 frames} (\cref{tab:s-eth50}). 13 scenes, official frames (one frame set, so no seed std), one locked pass per configuration. The exact engine and \ra stay within 0.6 AUC@3 of Official; \raf loses 1.65 against the exact engine and 2.22 against Official ($-1.14$ at $N{=}10$).
\begin{table}[t]
\centering\small\setlength{\tabcolsep}{3pt}
\caption{ETH3D at $N{=}50$ (13 scenes, official frames).}
\label{tab:s-eth50}
\begin{tabular}{@{}lcccc@{}}
\toprule
 & AUC@3 & AUC@30 & $\Delta_3$ vs Exact & $\Delta_3$ vs Off. \\
\midrule
Official \vggto & 66.13 & 95.13 & $+0.57$ & 0 \\
Exact engine & 65.56 & 95.05 & 0 & $-0.57$ \\
\ra & 65.62 & 95.06 & $+0.06$ & $-0.51$ \\
\raf & 63.91 & 94.82 & $-1.65$ & $-2.22$ \\
threshold schedule & 65.75 & 95.08 & $+0.19$ & $-0.38$ \\
sensitivity-only greedy (post hoc) & 65.25 & 95.03 & $-0.30$ & $-0.88$ \\
3-sample, first run & 64.65 & 94.93 & $-0.90$ & $-1.48$ \\
3-sample, seed-matched & 65.03 & 94.98 & $-0.53$ & $-1.11$ \\
\bottomrule
\end{tabular}
\end{table}

\paragraph{Depth and point maps} (\cref{tab:s-depth}). Depth follows \vggto's \texttt{depth\_metrics} unchanged for every method: each frame is aligned separately by a median scale and shift, valid ground truth in 0.1--100\,m. Point maps are reported for our engine only, because the Official \vggto runner does not compute them: predicted and ground-truth depth are unprojected with their own cameras and aligned by a Sim(3) (Umeyama, up to 20k seeded correspondences); this convention is ours, not \vggto's. Depth is unchanged; \ra matches the exact engine on point maps; \raf's Chamfer distance is 5\% (ETH3D) and 7\% (NRGBD) higher, in line with its small pose cost.
\begin{table}[t]
\centering\small\setlength{\tabcolsep}{2.2pt}
\caption{Depth ($\delta_{1.25}$ in \%, AbsRel) and point maps (Chamfer, GT units) at $N{=}10$. ETH3D: official frames (one value); 7-Scenes, NRGBD: three seeds.}
\label{tab:s-depth}
\resizebox{\linewidth}{!}{\begin{tabular}{@{}llccc@{}}
\toprule
 & & $\delta_{1.25}$ & AbsRel & Chamfer \\
\midrule
ETH3D & Official \vggto & 99.83 & 0.0115 & -- \\
 & Exact engine & 99.83 & 0.0115 & 0.0929 \\
 & \ra & 99.83 & 0.0116 & 0.0930 \\
 & \raf & 99.84 & 0.0117 & 0.0979 \\
\midrule
7-Scenes & Official \vggto & 94.61{\scriptsize$\pm$0.17} & 0.0520{\scriptsize$\pm$0.0010} & -- \\
 & Exact engine & 94.61{\scriptsize$\pm$0.17} & 0.0520{\scriptsize$\pm$0.0010} & 0.0395{\scriptsize$\pm$0.0014} \\
 & \ra & 94.60{\scriptsize$\pm$0.17} & 0.0520{\scriptsize$\pm$0.0010} & 0.0395{\scriptsize$\pm$0.0014} \\
 & \raf & 94.59{\scriptsize$\pm$0.19} & 0.0522{\scriptsize$\pm$0.0010} & 0.0396{\scriptsize$\pm$0.0015} \\
\midrule
NRGBD & Official \vggto & 99.67{\scriptsize$\pm$0.02} & 0.0072{\scriptsize$\pm$0.0001} & -- \\
 & Exact engine & 99.67{\scriptsize$\pm$0.02} & 0.0073{\scriptsize$\pm$0.0001} & 0.0221{\scriptsize$\pm$0.0020} \\
 & \ra & 99.67{\scriptsize$\pm$0.02} & 0.0073{\scriptsize$\pm$0.0001} & 0.0223{\scriptsize$\pm$0.0021} \\
 & \raf & 99.67{\scriptsize$\pm$0.02} & 0.0074{\scriptsize$\pm$0.0001} & 0.0237{\scriptsize$\pm$0.0026} \\
\bottomrule
\end{tabular}}
\end{table}

\paragraph{Timing spread} (\cref{tab:s-timing}). Each session is a fresh, locked process: 2 warm-ups, median of 5 runs. Values for the same configuration differ by at most 10\,ms across locked sessions. At $N{=}50$ the paper uses the mean of the three latest sessions for the exact engine and \ra, and for \raf the 1616.4\,ms session in which the rivals of Tab.~4 were timed (2.23\x); its three-session mean is 1611.8\,ms (2.24\x), and no other number in the paper uses that mean. At $N{=}200$ three later locked sessions give the exact engine $20748\pm8$, \ra $12515\pm8$, \raf $10557\pm4$ and the seed-matched 3-sample schedule $10656\pm5$\,ms; with one earlier session, the exact engine and \ra are $20745\pm9$ and $12512\pm9$\,ms, the values in the paper. Fourteen other locked, warmed-up the exact engine runs at $N{=}50$ since the start of this study have medians of 2227--2246\,ms ($2235\pm5$). An earlier table reported 2368\,ms for the exact engine; that value was the first, cold forward of an untimed accuracy pass and is not used.
\begin{table}[t]
\centering\small\setlength{\tabcolsep}{3pt}
\caption{Latency at $N{=}50$ (kitchen $512{\times}336$) over separate locked sessions: session medians and mean\,$\pm$\,std.}
\label{tab:s-timing}
\begin{tabular}{@{}lcc@{}}
\toprule
 & session medians (ms) & mean \\
\midrule
Official \vggto & 3605.0 & 3605 (1 session) \\
Exact engine & 2237.7, 2237.3, 2234.9 & $2236.6\pm1.5$ \\
\quad earlier & 2236.7, 2229.4 & \\
\ra & 1734.9, 1732.7, 1735.0 & $1734.2\pm1.3$ \\
\quad earlier & 1735.3, 1729.5 & \\
\raf & 1616.4, 1612.5, 1606.4 & $1611.8\pm5.1$ \\
3-sample, first run & 1616.0, 1616.4, 1615.7 & $1616.0\pm0.4$ \\
3-sample, seed-matched & 1612.9 & (1 session) \\
\bottomrule
\end{tabular}
\end{table}

\paragraph{Throughput at 2--10 frames} (\cref{tab:s-small}). At small $N$ global attention is a small share of the work: the exact engine alone is already 1.74/1.54/1.69\x faster than Official at $N{=}2/4/10$, and precision and schedule add only 1--9\%.
\begin{table}[t]
\centering\small\setlength{\tabcolsep}{3pt}
\caption{Latency in ms (frames per second) at small $N$, kitchen, one locked session. Peak memory: engine peak (ours) or \texttt{max\_memory\_allocated} with weights (Official), so the two are not strictly comparable.}
\label{tab:s-small}
\resizebox{\linewidth}{!}{\begin{tabular}{@{}lcccc@{}}
\toprule
 & $N{=}2$ & $N{=}4$ & $N{=}10$ & peak MiB ($N{=}2/4/10$) \\
\midrule
Official \vggto & 90.5 (22.1) & 167.5 (23.9) & 482.2 (20.7) & 6179 / 6265 / 6571 \\
Exact engine & 52.0 (38.5) & 108.5 (36.9) & 285.4 (35.0) & 2478 / 2766 / 3182 \\
\ra & 51.4 (38.9) & 105.4 (38.0) & 265.5 (37.7) & 2478 / 2766 / 3182 \\
\raf & 51.3 (39.0) & 104.7 (38.2) & 260.7 (38.4) & 2478 / 2766 / 3182 \\
\bottomrule
\end{tabular}}
\end{table}

\subsection{Protocol sensitivity: a three-sample search}
\label{sec:s-sens}
\raf was fixed by the single-sample protocol of the paper before any evaluation run; nothing below changed it. To test how much the schedule depends on that protocol, we re-ran the greedy search at \ra's precision with the same budget ($-1$ / $-0.5$) and visit order, accepting a step only if the cumulative configuration passes on all three calibration frame samples.
\emph{First run.} It accepts frame on 5 and 0 and merge$_4$ on 4, 1 and 19 (13 candidates). An audit found that, for seeds 1 and 2, its outdoor calibration cells reused the seed-0 frames. \emph{Seed-matched run.} With the correct frames, two of the first run's steps (frame on 0, merge$_4$ on 1) fail on seed 1, and the search returns frame on 5 and merge$_4$ on 0, 3, 4, 8 (13 candidates, 30 per-seed checks). Both schedules pass all three samples (\cref{tab:s-recheck}); the three-sample search is therefore itself sensitive to the calibration sample. Neither selects a matching layer.
\emph{Evaluation} (against the exact engine unless noted; worst over datasets per seed, mean\,$\pm$\,std over seeds):
\begin{itemize}\setlength\itemsep{0pt}
\item $N{=}10$, six datasets: first run $-0.84\pm0.34$, seed-matched $-0.75\pm0.19$, \raf $-0.80\pm0.17$; against Official $-1.16\pm0.65$, $-0.80\pm0.16$, $-1.14\pm0.00$; worst $\Delta$AUC@30 against Official $-0.71$, $-0.27$, $-1.13$.
\item $N{=}50$/100/200, five datasets: first run $-0.64\pm0.07$ / $-1.05\pm0.13$ / $-1.75\pm0.13$; seed-matched $-0.61\pm0.13$ / $-1.55\pm0.39$ / $-0.56\pm0.04$; \raf $-0.65\pm0.14$ / $-1.20\pm0.26$ / $-1.03\pm0.16$; threshold schedule $-0.62\pm0.21$ / $-1.87\pm0.35$ / $-1.50\pm0.24$. The weakest dataset at 100 frames is RE10K for every schedule; at 200 frames RE10K for the first run and the threshold schedule, NRGBD for \raf, 7-Scenes or RE10K for the seed-matched run. Against Official at $N{=}50$: seed-matched $-0.63\pm0.12$, first run $-0.80\pm0.10$, \raf $-0.95\pm0.13$.
\item ETH3D, $N{=}50$ (\cref{tab:s-eth50}): $-1.48$ and $-1.11$ against Official, \raf $-2.22$.
\item Latency, $N{=}50$: $1616.0\pm0.4$ (3 sessions) and 1612.9\,ms (1 session), \raf 1611.8\,ms mean; $N{=}200$: first run 10.61\,s (1 session), seed-matched $10.66\pm0.01$\,s and \raf $10.56\pm0.00$\,s (3 sessions each).
\end{itemize}
The first run is similar to \raf up to 100 frames and loses more at 200 ($-1.75$ on RE10K vs $-1.03$ on NRGBD). The seed-matched schedule is the most accurate searched schedule at 10, 50 and 200 frames and on ETH3D at 50 frames, at 0.9\% more latency than \raf at 200 frames, but the least accurate searched schedule at 100 frames ($-1.55$, RE10K on every seed). No schedule dominates; the gaps (at most 1.2 AUC@3) are comparable to the seed std at 100 frames, and the choice among them depends on the calibration sample.

\paragraph{The AVGGT/GPA-VGGT layer split in our engine.} The in-engine control of Tab.~4 of the paper runs frame attention on blocks 0, 1, 3, 4, 5, 7, 8 and $2{\times}2$ mean merging with $+\log c$ on blocks 10--13, 15--19, 21--23, at \ra's precision. Its layer split equals that of GPA-VGGT (frame on the non-register blocks of 0--8, merging on 9--23, of which 9, 14 and 20 are register-only) and AVGGT's published ranges, but the merge keeps a quarter of each non-reference frame's patch tokens, whereas GPA-VGGT keeps half ($r{=}0.5$) with its own token selection; it is a layer-choice control, not GPA-VGGT. Worst $\Delta$AUC@3 against Official: $-37.89$ on all three seeds (ETH3D), worst $\Delta$AUC@30 $-28.41\pm3.04$; 1360\,ms at $N{=}50$ (1.6\x the exact engine's speed), 6.63\,s at $N{=}200$.

\paragraph{Reference frame: position or content?} We ran the \vggto trunk (PyTorch, bf16, released weights) on four 10-frame sequences (ETH3D courtyard and electro, Mip-NeRF~360 room and counter) in the original order and with the middle frame moved to the first slot, and compared the attention mass on the frame in slot 0. In the anchoring blocks it is unchanged: 0.316/0.315 (block 15), 0.632/0.635 (19), 0.610/0.605 (21), 0.528/0.527 (23), with $|\Delta|\le0.005$, smaller than the std of $\Delta$ over the sequences (0.012--0.024); diffuse and matching blocks change by at most 0.013. Anchoring follows the position of the reference frame, not its content. (A fifth, 50-frame 7-Scenes cell failed to load and is not included.)

\section{Layer-placement and routing controls}
\label{sec:s-ctrl}
All runs: RTX 4070 SUPER, GPU lock held; accuracy at $N{=}10$ on the six evaluation datasets, seeds 0, 1, 2, $\Delta$ against the exact engine, worst = per-seed minimum over datasets, mean\,$\pm$\,std (ddof${=}1$). Latency: $N{=}50$, one locked session in which the exact engine, \ra and \raf took 2230, 1732 and 1610\,ms. These controls were run after the main evaluation.
\paragraph{Random placement.} Five draws place \raf's operator counts (three frame, two merge$_4$ layers) at random, at the precision of \ra. \emph{Non-matching}: drawn from the 13 non-register blocks outside 10--16. \emph{Regime operator space}: frame drawn from context blocks \{0, 1, 3, 4, 5, 7, 8\}, merge$_4$ from context or anchoring blocks \{0--8, 17--23\} (non-register).
\begin{table}[t]
\centering\small\setlength{\tabcolsep}{2.4pt}
\caption{Random placements of \raf's operators. Worst $\Delta_3$, worst $\Delta_{30}$ against the exact engine; ms at $N{=}50$.}
\label{tab:s-rand}
\resizebox{\linewidth}{!}{\begin{tabular}{@{}lccr@{}}
\toprule
schedule (frame; merge$_4$) & worst $\Delta_3$ & worst $\Delta_{30}$ & ms \\
\midrule
\ra & $-0.51${\scriptsize$\pm$0.11} & $-0.24${\scriptsize$\pm$0.11} & 1732 \\
\raf (0, 4, 5; 3, 19) & $-0.80${\scriptsize$\pm$0.17} & $-1.20${\scriptsize$\pm$0.00} & 1610 \\
\midrule
non-matching \#1 (0, 8, 23; 5, 18) & $-73.53${\scriptsize$\pm$2.48} & $-16.42${\scriptsize$\pm$1.06} & 1622 \\
non-matching \#2 (5, 8, 17; 7, 23) & $-59.37${\scriptsize$\pm$1.81} & $-18.99${\scriptsize$\pm$2.40} & 1624 \\
non-matching \#3 (4, 18, 19; 3, 5) & $-83.18${\scriptsize$\pm$1.35} & $-50.59${\scriptsize$\pm$0.00} & 1612 \\
non-matching \#4 (1, 3, 19; 5, 18) & $-82.25${\scriptsize$\pm$1.85} & $-40.28${\scriptsize$\pm$0.00} & 1623 \\
non-matching \#5 (3, 19, 22; 1, 5) & $-82.50${\scriptsize$\pm$1.20} & $-40.58${\scriptsize$\pm$0.00} & 1621 \\
\midrule
regime \#1 (0, 1, 5; 4, 17) & $-4.14${\scriptsize$\pm$1.13} & $-0.91${\scriptsize$\pm$0.10} & 1613 \\
regime \#2 (4, 7, 8; 5, 19) & $-1.45${\scriptsize$\pm$0.70} & $-0.45${\scriptsize$\pm$0.28} & 1612 \\
regime \#3 (0, 4, 8; 19, 23) & $-2.64${\scriptsize$\pm$0.76} & $-0.50${\scriptsize$\pm$0.23} & 1614 \\
regime \#4 (0, 4, 5; 8, 17) & $-4.23${\scriptsize$\pm$0.76} & $-1.48${\scriptsize$\pm$0.15} & 1611 \\
regime \#5 (0, 3, 5; 1, 23) & $-2.99${\scriptsize$\pm$1.11} & $-0.52${\scriptsize$\pm$0.29} & 1626 \\
\bottomrule
\end{tabular}}
\end{table}
Every non-matching draw collapses (mean $-76.2$; NRGBD worst on every seed) because frame attention lands on a late block (17, 18, 19, 22 or 23), which the single-layer map already gives $-48$ to $-77$ calibration AUC@3. Within the regime operator space the loss falls to 1.5--4.2 (mean $-3.1$), still 1.8--5.3\x \raf's at the same latency; merging on 17 or 23 costs most. The operator-per-regime assignment thus explains almost all of the gap, and the calibration check a further 0.6--3.4 points.
\paragraph{Sensitivity-only greedy.} All 38 (block, operator) pairs on the 19 non-register blocks are visited in order of their own single-layer calibration cost (worst $\Delta_3$, then worst $\Delta_{30}$, then mean $\Delta_3$), with no regime statistic and no preference for frame attention. A pair is accepted if the cumulative configuration stays within the operator budget of Sec.~4.4 of the paper ($-1$ / $-0.5$, same calibration sample as \raf); pairs that fail on their own are skipped (22 pairs: every pair on blocks 10--18, 21, 23, and frame on 19 and 22). Steps (calibration worst $\Delta_3$ / $\Delta_{30}$): +0:merge$_4$ $-0.02/-0.03$; +8:merge$_4$ $-0.50/-0.08$; +3:merge$_4$ $-0.54/-0.14$; +19:merge$_4$ $-0.35/-0.04$; +5:frame $-0.33/-0.14$; +4:merge$_4$ $-0.75/-0.17$; 7:merge$_4$ $-1.14$ (rejected); 7:frame $-1.08$ (rejected); +1:merge$_4$ $-0.74/-0.23$; +22:merge$_4$ $-0.99/-0.20$. 48 configurations (38 reused single-layer evaluations, 10 cumulative trials). The result, merge$_4$ on 0, 1, 3, 4, 8, 19, 22 and frame on 5, differs from threshold schedule only on blocks 5 (frame instead of merge$_4$) and 7 (dense). On evaluation it scores worst $\Delta_3$ $-0.91\pm0.25$ and $\Delta_{30}$ $-0.21\pm0.19$ (per dataset: ETH3D $+0.40$, CO3D $-0.27$, RE10K $-0.61$, 7-Scenes $-0.51$, NRGBD $-0.03$, TUM $-0.40$) at 1558\,ms ($N{=}50$) and 9.67\,s ($N{=}200$; the exact engine 20.76\,s and \raf 10.56\,s in the same session; peak 9.7\,GiB). At $N{=}10$ it ties \raf on worst AUC@3 and is better on AUC@30. \emph{Long sequences} (five datasets, three seeds, \cref{tab:s-longN}): worst $\Delta_3$ $-0.81\pm0.43$ / $-3.42\pm0.46$ / $-2.10\pm0.14$ at 50/100/200 frames (\raf $-0.65$ / $-1.20$ / $-1.03$), worst $\Delta_{30}$ $-0.11$ / $-0.74$ / $-0.34$; the loss is RE10K on every seed (per seed $-3.16$, $-3.14$, $-3.95$ at 100; $-2.24$, $-1.97$, $-2.10$ at 200), the other four datasets stay within $\pm0.25$. It is worse there than the threshold schedule it nearly equals ($-1.87$/$-1.50$). On ETH3D at 50 frames it scores $-0.30$ against the exact engine and $-0.88$ against Official (\raf $-1.65$/$-2.22$). On the three-sample calibration check it passes 2/3 (seed 1 fails on the indoor dataset, $-1.26$/$-0.74$), the same dataset that fails at long $N$. Latency: 1556.8$\pm$0.2\,ms over three further locked sessions (\raf 1611.6, 3.5\% slower) and 9.67\,s at $N{=}200$ in a second session (\raf 10.55\,s). We therefore keep \raf as the named point; the sensitivity ranking finds cheap layers, but the $N{=}10$ single-sample protocol cannot see long-$N$ failures, which the three-sample check flags.
\paragraph{Top-$k$ frame routing at 200 frames.} No routing kernel exists in our engine, so this control runs the official \vggto in PyTorch (pose only; bf16 aggregator weights; frame blocks, patch embedding and global MLPs chunked, which is exact). With exact attention this harness reproduces the exact engine's $N{=}200$ AUC@3 within seed noise (7-Scenes 26.29 vs 26.3, NRGBD 86.62 vs 86.3, TUM 41.77 vs 42.0, CO3D 58.39 vs 58.8, RE10K 50.90 vs 50.7). In \emph{every} global block, each query frame attends to the keys of $k{=}\lceil fN\rceil$ frames: its own, frame 0 and the $k{-}2$ frames with the highest router score, either head-averaged $\langle\bar q_f,\bar k_g\rangle$ over patch tokens (pooled) or the same on camera tokens (cam, BlazeVGGT-style). Five datasets (ETH3D has fewer than 200 frames per scene), 6 sequences each. Worst $\Delta$AUC@3 / $\Delta$AUC@30 against the same harness with exact attention: top-50\% cam $-17.07\pm1.10$ / $-3.18\pm0.17$ (3 seeds); top-50\% pooled $-49.94$ / $-13.82$; top-25\% cam $-36.72$ / $-6.34$; top-25\% pooled $-74.57$ / $-25.96$ (seed 0). NRGBD is worst in every arm. In this gather-based PyTorch implementation, 50\% routing is not faster than exact attention (24.1 vs 22.8\,s) and 25\% is 1.48\x faster (15.4\,s). The camera-token router is consistently better than pooled features, so selection is not meaningless, but even the better half of the frames misses mass the matching layers need. This is a generic port, not BlazeVGGT's or Spark3R's own method, which report their gains at 1,000--2,000 frames with their own layer choices; routing restricted to context layers was not tested.

\section{Implementation details}
Engine: native CUDA engine, NVRTC run-time compilation (compute\_89), cuBLASLt for bf16 GEMMs. Weights: bf16 arena (FP8 companion only for FP8 blocks, with per-channel scales). Attention kernels: INT8 \texttt{mma.m16n8k32} for $QK^\top$ with per-16-row $Q$ and per-64-key $K$ scales after key smoothing; FP8 E4M3 \texttt{mma} for $PV$ with $V$ scaled to $\pm2.25$ (FP16 accumulation); 16-bit $PV$ with $V$ scaled to $\pm256$ and FP16 accumulation per 64-key block folded into FP32; log-count bias added to the INT32 scores in registers. Sage workspaces borrow the idle MLP buffer. CUDA graphs for $N\le32$. Timing: CUDA events around trunk + camera + dense heads, pre-loaded images, median after 2 warm-ups, exclusive \texttt{flock} on the GPU.

\section{Reproducibility checklist}
\begin{itemize}\setlength\itemsep{0pt}
\item \textbf{Weights}: released \vggto-1B checkpoint, unchanged; no training or fine-tuning.
\item \textbf{Data}: public datasets only; calibration and evaluation data in \cref{tab:s-split}; evaluation manifests (frame maps per seed) released with the code.
\item \textbf{Seeds}: three frame-sampling seeds for Tabs.~1, 4 and 5 of the paper; seed 0 where stated. The engine is deterministic (bit-identical re-runs).
\item \textbf{Metrics}: the released \vggto \texttt{eval/metrics.py} (pose AUC, depth with median alignment); Inria 3DGS metrics for the NVS table of the supplement.
\item \textbf{Hardware}: RTX 4070 SUPER 12\,GB (CUDA 12.8) and RTX PRO 6000 Blackwell 96\,GB (CUDA 13.0); every timing under an exclusive lock or on an exclusive card.
\item \textbf{Baselines}: rival code at pinned commits (GPA-VGGT 767b172, sparse-vggt 7e8824f, FastVGGT 6526e27, SpargeAttn a2666ad), ported onto \vggto by monkey-patching attention; no $\Omega$ configuration was released by the authors.
\item \textbf{Logs}: every number has a record (experiment id, config, commit, GPU state, lock flag) in our results files.
\item \textbf{Code}: engine and scripts will be released.
\end{itemize}

\end{document}